\documentclass[10pt,twocolumn,letterpaper]{article}

\usepackage{wacv}              

\usepackage{graphicx}
\usepackage{amsmath}
\usepackage{amssymb}
\usepackage{booktabs}

\usepackage[accsupp]{axessibility}  

\usepackage{amsmath,amsfonts,bm}

\def\eqref#1{equation~\ref{#1}}

\def\ceil#1{\lceil #1 \rceil}

\def\1{\bm{1}}

\DeclareMathAlphabet{\mathsfit}{\encodingdefault}{\sfdefault}{m}{sl}
\SetMathAlphabet{\mathsfit}{bold}{\encodingdefault}{\sfdefault}{bx}{n}

\newcommand{\R}{\mathbb{R}}

\newcommand{\KL}{D_{\mathrm{KL}}}

\newcommand{\normlone}{L^1}

\usepackage[pagebackref,breaklinks,colorlinks]{hyperref}

\usepackage[capitalize]{cleveref}
\crefname{section}{Sec.}{Secs.}
\Crefname{section}{Section}{Sections}
\Crefname{table}{Table}{Tables}
\crefname{table}{Tab.}{Tabs.}

\usepackage{graphicx}
\usepackage{multirow}
\usepackage{algorithm}
\usepackage{algpseudocode}
\usepackage{float}
\usepackage{booktabs}
\usepackage{comment}
\usepackage{cuted}
\usepackage{rotating}
\usepackage{pdflscape}
\usepackage{enumitem}
\usepackage{url}
\usepackage{tikz}

\usepackage{xcolor}
\usepackage{pifont}

\newcommand{\partialft}[0]{\textsc{Partial}}
\newcommand{\linear}[0]{\textsc{Linear}}
\newcommand{\fullft}[0]{\textsc{Full}}
\newcommand{\sidetune}[0]{\textsc{Sidetune}}
\newcommand{\mlp}[0]{\textsc{Mlp}}
\newcommand{\bias}[0]{\textsc{Bias}}
\newcommand{\adapter}[0]{\textsc{Adapter}}

\def\ie{\emph{i.e.}}

\newcommand{\cub}[0]{CUB-200-2011}
\newcommand{\nabirds}[0]{NABirds}
\newcommand{\flowers}[0]{Oxford Flowers}
\newcommand{\cars}[0]{Stanford Cars}
\newcommand{\dogs}[0]{Stanford Dogs}
\newcommand{\vtab}[0]{VTAB-1k}

\newcommand{\method}[0]{MiMi}
\newcommand{\spacefig}{\vspace{-1mm}}

\newcommand\imad[1]{\textcolor{black}{#1}}

 \newcommand\review[1]{\textcolor{black}{#1}}

\begin{document}

\title{Mini but Mighty: Finetuning ViTs with Mini Adapters}

\author{Imad Eddine Marouf
\qquad
Enzo Tartaglione
\qquad
Stéphane Lathuilière\\
LTCI, Télécom-Paris, Institut Polytechnique de Paris, France\\
{\tt\small imad.marouf@ip-paris.fr}
}
\maketitle

\begin{abstract}
   Vision Transformers (ViTs) have become one of the dominant architectures in computer vision, and pre-trained ViT models are commonly adapted to new tasks via finetuning. Recent works proposed several parameter-efficient transfer learning methods, such as adapters, to avoid the prohibitive training and storage cost of finetuning. 

    In this work, we observe that adapters perform poorly when the dimension of adapters is small, and we propose \method, a training framework that addresses this issue. We start with large adapters which can reach high performance, and iteratively reduce their size. 
    To enable automatic estimation of the hidden dimension of every adapter, we also introduce a new scoring function, specifically designed for adapters, that compares the neuron importance across layers. Our method outperforms existing methods in finding the best trade-off between accuracy and trained parameters across the three dataset benchmarks DomainNet, VTAB, and Multi-task, for a total of 29 datasets.\footnote{Code is available: \url{https://github.com/IemProg/MiMi}}
\end{abstract}

\section{Introduction}

Transformers have gained increasing attention owing to their outstanding performance~\cite{ViT, Swin, DeiT, vaswani2017attention}: Vision Transformers (ViTs) trained on large-scale datasets have demonstrated a remarkable ability to learn new tasks~\cite{ViT}. The most commonly adopted strategy to learn new tasks consists of fully or partially fine-tuning a pre-trained network; however, when dealing with multiple tasks, this approach necessitates training multiple separate models, which results in large storage costs.
\begin{figure}[t]
    \begin{center}
    \includegraphics[width=1\linewidth]{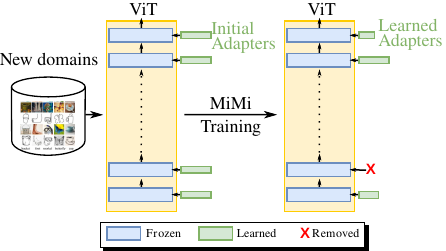} 
    \end{center}
    \caption{Layer-wise small blocks are injected into ViTs to efficiently adapt to new domains. \method{} estimates the best rank for each adapter weight, it reduces the number of parameters and removes completely injected adapters for some layers if necessary.}
    \label{fig:teaser}
     \spacefig
\end{figure}

Recently, Parameter-Efficient Training (PET) approaches have been developed to help large pre-trained models adapt to new tasks, with minimal added parameters~\cite{MaMAdapter, LoRa, VPT}. Among these, adapters~\cite{EfficientNLP} and its variants~\cite{MaMAdapter, Compacter, AdaptersHyperNetworks} are frequently employed for Natural Language Processing (NLP) tasks. Adapters are small modules inserted into transformer blocks, which enable efficient adjustment of the data representation to the downstream task: they offer similar performance to full fine-tuning (\ie~updating all parameters) while requiring a very low number of trainable parameters~\cite{EfficientNLP, AdapterFusion}.

When it comes to vision tasks, PET approaches are mostly explored for convolutional neural networks~\cite{BudgetAware, PiggyBack, RebuffiBV17, RebuffiBV18}. In contrast, several PET approaches have been proposed in NLP tasks: here, adapters are Multi-Layer-Perceptrons (MLPs) equipped with residual connections~\cite{EfficientNLP}. These multi-layer adapters can fit new tasks with enough representation power and the size of their hidden layers provides a simple trade-off between performance and parameter efficiency~\cite{EfficientNLP}. Nevertheless, they suffer from two weaknesses. First, the performance drops when the size of multi-layer adapters is too small~\cite{AdapterFormer} (as confirmed by our experiments -see Se.~\ref{sec:main_results}, and supplementary material-). Second, the optimal hyper-parametrization of 
adapters is complex: the hidden layer dimensions must be specified for every adapter in every layer, and its optimal size 
depends on the downstream task. Thus, these adapters cannot be employed where the available storage is limited.

In this work, we propose a training scheme named \method{} (Fig.~\ref{fig:teaser}) which addresses these two limitations. 
Our approach facilitates efficient parameter allocation by predominantly assigning additional parameters to layers that genuinely necessitate adaptation to the new task (Fig.~\ref{fig:globa_vs_local}). 
More specifically, we start by training adapters with high-dimensional hidden spaces and gradually decrease their dimensionality by identifying neurons that can be omitted in each adapter. Additionally, we introduce a novel scoring criterion to determine the layers where more adaptation is needed, 
which enables the comparison of a ``neuron importance'' among adapters in various layers.

\noindent Our work makes the following key contributions:
\begin{itemize}[noitemsep,nolistsep]
    \item We propose a novel iterative training scheme for learning small adapters for ViTs.
    \item We present a new scoring function that can effectively compare the significance of neurons across adapters. This approach enables us to estimate the optimal hidden dimension of adapters for ViTs automatically, which leads to a more efficient parameter allocation. 
    \item Finally, we compare the proposed approach with multiple PET methods designed for both NLP and vision tasks using a total of 29 datasets. From these experiments, we draw several conclusions: (i) we demonstrate that our approach obtains the best performance in terms of accuracy among methods with similar numbers of parameters; (ii) our ablation study validates the positive impact of our adaptive strategy to automatically estimate the hidden dimension of adapters. 
\end{itemize}

\section{Related Work}
\label{related_work}

\paragraph{Vision Transformers.} 
Originally designed for NLP tasks, Transformers~\cite{VaswaniSPUJGKP17} have recently been adapted for vision tasks, such as image classification. Vision Transformers (ViTs) divide the image into patches, process them as token embeddings, and employ transformer encoders with self-attention to learn image representations~\cite{ViT}. ViTs have shown impressive performance, outperforming ConvNets in some cases~\cite{CNNmeetsViT}. However, their large parameter count leads to significant storage costs, limiting complete finetuning for each new task. This motivates our study. 
While Swin~\cite{Swin} is a widely adopted ViT due to its excellent performance across vision tasks, our approach of using tiny adapters can be applied to any ViT architecture (see Sec.~\ref{BackboneViTs}).\\

\noindent\textbf{Network Pruning.}
When referred to deep neural networks, pruning consists of reducing the number of parameters of a pre-trained model~\cite{LeCun, SongPrune}. It can be roughly categorized into two groups: (i) unstructured pruning, which removes the least significant weights (according to certain criteria like weight magnitude~\cite{SongHanMagnitudePrune} or gradient magnitude~\cite{GradientPrune}) without a specific structure to be followed; (ii) structured pruning, which focuses in removing model sub-structures, like channels~\cite{EnzoPaper, HeZS17} or attention heads~\cite{HeadsPrune}. 
Pruning techniques usually reduce the number of parameters in a network trained for a specific task, while \method{} decreases the number of parameters added through adapters that fit the model to a new task without altering the original model's parameters.
SparseAdapters (SA)~\cite{SparseAdapters}, show that applying unstructured pruning to adapters~\cite{EfficientNLP} achieves comparable performance. In comparison to SA, our method incorporates a look-ahead strategy that considers the effects of upsampling layers, while SA does not. Furthermore, MiMi employs structured pruning, whereas SA utilizes unstructured pruning techniques to reduce the size of adapters and remove them if necessary.
\imad{GraSP~\cite{GraSP} utilizes Hessian-gradient products for each layer, discarding weights with elevated scores in a single move, emphasizing those that improve gradient flow. Conversely, SNIP~\cite{SNIP} determines layer gradients using sampled mini-batch data, assigning scores and eliminating weights with the highest scores in one step. However, both these approaches are not apt for pruning Adapters. Our experiments show that they do not perform well when applied to adapters.}

\noindent\textbf{Efficient Transformers Finetuning.}
ViTs' lack of typical CNN inductive biases makes their finetuning on new tasks susceptible to overfitting~\cite{CNNmeetsViT, ViTSmallDsets}. Additionally, the need to update all the parameters and store a separate model copy per task hinders scalability and real-world applicability. To tackle this, three types of approaches have emerged: (i) updating only newly added parameters~\cite{AdapterFormer, EfficientNLP, LoRa, VPT, AdapterFusion}; (ii) sparse parameter updates~\cite{BitFit, LoRa}; and (iii) low-rank factorization for weight updates~\cite{Compacter}. While prompt techniques like \textit{VPT}~\cite{VPT} achieve excellent performance, they lack flexibility for downstream tasks that differ significantly from pre-training~\cite{AdapterFormer}.

Our work falls into the first category, building on adapters~\cite{EfficientNLP} for NLP tasks. However, we introduce a specific training algorithm enabling high performance, with small-size adapters for downstream tasks. Unlike previous adapter approaches~\cite{AdapterFormer, EfficientNLP} with fixed-size adapters across layers~\cite{AdapterFormer}, \method{} dynamically assesses adapter sizes and even removes them if necessary. By minimizing trainable parameters, \method{} enhances performance and reduces storage footprint in multi-task scenarios. Our preliminary results demonstrate that different layers require different adapter sizes, as highlighted in the supplementary material.\\
In contrast, \cite{autopeft, SPET} utilize Neural Architecture Search (NAS) to identify optimal PET configurations, facilitating an expansive configuration search space populated by various representative PET methods. However, this approach is notably computation-intensive and uses different PET modules. Our work is primarily concentrated on determining the appropriate adapter size for each layer and eliminating certain adapters when deemed essential.

\section{Proposed Method}
\label{method}
In this section, we start with the description of adapters~\cite{EfficientNLP} and discuss their practical benefits. Then, we introduce \method, our method to estimate the hidden dimension for each adapter that can effectively maintain high performance with fewer parameters for memory efficiency.

\subsection{Preliminaries}
Our objective is to adapt a pre-trained ViT network for a new task by incorporating small modules called ``adapters'' into the existing layers. This adaptation process involves training the linear classifier (referred to as the ``head'') and the adapter parameters while keeping the weights of the original model frozen. In our training procedure, we focus on describing the adapter parameters, and 
he linear classifier parameters are also learned simultaneously.

\begin{figure}
\centering
  \includegraphics[width=\columnwidth]{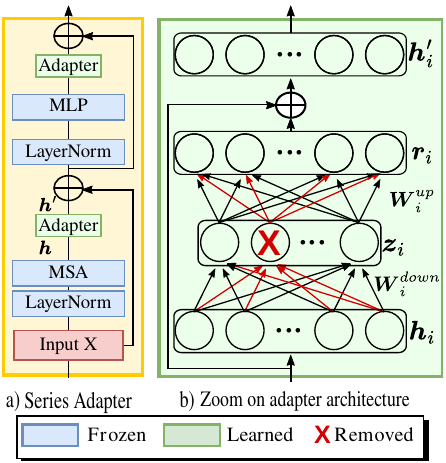}
\caption{The adapter structure injected into ViT model (a), and our approach to adjust the adapter's size (b). MSA and MLP are multi-head self-attention and feed-forward blocks, respectively.}
\label{fig:main}
\end{figure}

ViT architectures, such as the original ViT~\cite{ViT} or Swin~\cite{Swin}, consist of layers with two main sub-layers: a multi-head self-attention (MSA) layer and a fully-connected layer (MLP). Layer normalization is applied before each of these sub-layers, and residual connections are employed to skip MSA and MLP. In our approach, we introduce two adapters after each sub-layer. The adapter is directly applied to the output of the corresponding sub-layer, as depicted in Fig.~\ref{fig:main}a. The internal structure of the adapters is illustrated in Fig.~\ref{fig:main}b.

Considering the $i$-th adapter added to our pre-trained ViT, let $\boldsymbol{h}_i\!\in\!\R^{M_i}$ denote its input, of size $M_i$. 
Following \cite{EfficientNLP}, adapters employ a first fully-connected layer that down-projects $\boldsymbol{h}_i$ into $\boldsymbol{z}_i \in \R^{N_i}$ with some non-linear activation $\phi(\cdot)$. This layer is parametrized by a linear projection matrix $\boldsymbol{W}^{\text{down}}_i\in \R^{M_i\times N_i}$. Then, a second fully connected layer with parameters $\boldsymbol{W}^{\text{up}}_i\in \R^{N_i\times M_i}$ up-samples $\boldsymbol{z}_i$, producing as output $\boldsymbol{r}_i\in \R^{M_{i}}$. Finally, a residual skip-connection is employed inside the adapter module such that, if $\boldsymbol{r}_i$ is close to zero, 
the whole adapter module degenerates to an identity function. To summarize, given the input vector $\boldsymbol{h}_i$, the output vector $\boldsymbol{h}_i'$ is calculated as:
\begin{equation}
    \boldsymbol{h'}_i = \boldsymbol{W}^{\text{up}}_i\cdot \phi\left( \boldsymbol{W}^{\text{down}}_i\cdot \boldsymbol{h}_i \right) + \boldsymbol{h}_i.
    \label{eq1}
\end{equation}
The total number of parameters in the adapter is equal to $2\!\cdot\!N_i\!\cdot\!M_i$: since $M_i$ is fixed, we generally choose $N_i$ such that $N_{i} \ll M_{i}$ to obtain a low number of parameters. We define the compression rate $\sigma_i$ of an adapter as $\sigma_i\!=\!\frac{M_i}{N_i}$.

Previous works~\cite{EfficientNLP, AdapterFusion, AdapterDrop} have employed adapters with a uniform hidden dimension $N_i$ for all the adapters. However, this approach may not be optimal as early and late layers to the input of the model may focus on different types of patterns~\cite{AdapterFormer, IdentityCrisis} (see supplementary material). If we enable dynamic adjustment of the adapter's hidden dimension $N_i$ (or equivalently, $\sigma_i$) and determine their injection points, we enhance adaptation to downstream tasks effectively.

\subsection{Overview of \method}
\label{overall_method}

Let $\boldsymbol{W}^{\text{ViT}}$ be the initial parameters of the ViT model which are frozen through the whole adaptation process. With \method~our goal is to learn $\boldsymbol{W}^{\text{ada}}$, the set containing the adapter parameters $\boldsymbol{W}_i^{\text{ada}} = \boldsymbol{W}_i^{\text{up}} \cup \boldsymbol{W}_i^{\text{down}}$ of every $i$-th ViT sub-layer. In previous works~\cite{EfficientNLP,RebuffiBV18} $\boldsymbol{W}_i^{\text{ada}}$ is straightforwardly learned with stochastic gradient descent-based optimization; however, in our experiments (see Sec.~\ref{results}) we show that this approach does not perform well in the case of tiny adapters (small $N_i$ values). We start from the observation that, with the existing optimizers, sequentially training and pruning a large network is a successful strategy to find small networks with good performance, while directly training small networks usually suffers from optimization issues~\cite{frankle2018lottery}.
Therefore, we propose to start from large adapters and adopt an iterative pruning strategy that iteratively reduces their dimensions as detailed in Alg.~\ref{alg:ourmethodsimplified}.
\begin{algorithm}[t]
\caption{\method}
\label{alg:ourmethodsimplified}
\begin{algorithmic}[1]
\Procedure{\method ($\boldsymbol{W}^{ViT}$, $\boldsymbol{W}^{ada}$, $\rho$, $\sigma^{target}$)}{}
    \State Learn $\boldsymbol{W}^{ada}$  \label{line:vantrain}\Comment{$\boldsymbol{W}^{\text{ViT}}$ is frozen}
    \While{$\sigma < \sigma^{\text{target}}$}
        \State Sort $\boldsymbol{W}^{\text{ada}}$ according to $\mathcal{I}^{ij}$ (Sec.~\ref{sec:score})
        \State $\boldsymbol{W}^{\text{ada}} \gets$ top$(1\!-\!\rho)$  in $\boldsymbol{W}^{\text{ada}}$ \Comment \review{Selection} \label{line:selection1}
        \State Fine-tune $\boldsymbol{W}^{\text{ada}}$ \label{line:retrain}\Comment{$\boldsymbol{W}^{\text{ViT}}$ is frozen}
    \EndWhile
    \State \Return $\boldsymbol{W}^{\text{ada}}$ 
\EndProcedure

\end{algorithmic}
\end{algorithm}
We initialize every adapter with a hidden dimension proportional to its input dimension. We start from compression rates $\sigma_i\!=\!\sigma_0$ for every layer, where $\sigma_0$ is the initial compression rate. 
In our first training stage (line~\ref{line:vantrain}), we learn the adapter parameters $\boldsymbol{W}^{\text{ada}}$ via cross-entropy loss minimization using stochastic gradient descent. Then, we estimate a score that measures the importance of each adapter's neurons (more details will be provided in Sec.~\ref{sec:score}). This score is used to select the \textbf{neurons} that have the smallest impact on the adapter outputs; more precisely, we remove the bottom fraction $\rho$ of neurons from 
$\boldsymbol{W}^{\text{ada}}$ (line~\ref{line:selection1}). The remaining ones will constitute the new adapter configuration, and the hidden space sizes $N_i$ are updated accordingly. 
If the achieved average compression rate $\sigma$ is still lower than the target $\sigma^{\text{target}}$, another compression iteration follows; otherwise, the achieved configuration will be returned and the method stops. Note that the total number of training cycles \review{$C$} is given by:
\begin{equation}
    C = \left\lceil\frac{\log\left(\sigma^{0}\right)-\log\left(\sigma^{\text{target}}\right)}{\log\left(\rho\right)}-1\right\rceil,
\end{equation}
where $\ceil{\cdot}$ denotes the ceiling function. Therefore, our training scheme stops after a deterministic number of iterations that can be computed in advance. 
While we employ a stopping criterion based on a specific target compression rate, a target performance on a validation set could also be used.

\subsection{Importance Score in \method{}}
\label{sec:score}
In this section, we present the importance score function that we use in our training algorithm. Our design of the scoring function is motivated by the observation that, if an entire row in $\boldsymbol{W}_{i}^{\text{down}}$ and an entire column in $\boldsymbol{W}_{i}^{\text{up}}$ are equal to zero, then our adapter is strictly equivalent to one with \imad{a smaller dimension $M_i$}. Therefore, we propose a novel scoring function to employ the sum of the $\normlone$ norm of the corresponding row in $\boldsymbol{W}_{i}^{\text{down}}$ and the corresponding column in $\boldsymbol{W}_{i}^{\text{up}}$. More precisely, our importance score is formulated as follows:
\begin{equation}
    \label{eq:impscore}
    \mathcal{I}^{ij} = \frac{1}{N_i + M_i}\left( \sum_{k=1}^{M_i}\Big|W_{i}^{\text{down}}[j,k]\Big| + \sum_{k=1}^{N_i}\Big|W_{i}^{\text{up}}[k,j]\Big|\right),
\end{equation}
where $[\cdot,\cdot]$ denotes the matrix indexing operator. This importance score can be interpreted as a ``look-ahead'' strategy, where we observe, besides the output of a specific $j$-th neuron in the hidden space, also the impact of such an output in the next layer. 
Note that this formulation is based only on \textit{the magnitude of parameters belonging to the same neuron} of down-sampling, and its corresponding up-sampling one, and not on the magnitude of activations. This makes the importance score more computationally efficient since activation-based scoring depends on the input images, and consequently, statistics should be gathered at the batch or the dataset level, inducing non-negligible computation overhead. Furthermore, this choice is empirically supported by many works in the literature, like~\cite{chauvin1988back, SongHanMagnitudePrune, Li2017PruningFF, Renda2020Comparing}. A noteworthy element is that $\mathcal{I}^{ij}$ is normalized by the total number of parameters associated with a specific dimension of the adapter: this enables fair comparison across adapters, despite different input and hidden layer sizes. More details behind the motivation of our choice for \eqref{eq:impscore} are provided in the supplementary material.

\section{Experiments}
\label{results}

\label{experiments_setup}
We provide here the details about the datasets and our experimental setup.

\noindent \textbf{Datasets.} We evaluate our methods using the protocol adopted by \cite{ViTSmallDsets}, which consists of ten datasets for image classification tasks divided into two benchmarks. The first benchmark is known as \emph{DomainNet}~\cite{DomainNet}. It contains six different visual domains, which makes the finetuning experiments non-trivial. Since \emph{DomainNet} does not have a labeled testing set, we use the validation dataset for testing, as in \cite{DomainNet}. 
The second benchmark contains CIFAR-10/CIFAR-100~\cite{CIFAR100}, Oxford Flowers102~\cite{Nilsback08}, and SVHN~\cite{37648}, which are widely used as low-regime training datasets. Contrarily to DomainNet, these datasets are not single-task oriented but contain a larger variety of domains/tasks. We refer to them as belonging to the \emph{Multi-task} benchmark. Additionally, we provide an evaluation of the \emph{VTAB} benchmark~\cite{zhai2019vtab}, consisting of 19 diverse visual tasks (see supplementary).

\noindent\textbf{Implementation Details.} We follow the training protocol adopted by \cite{ViTSmallDsets}. We conduct our experiments with the official pre-trained Swin-T~\cite{Swin} ($\sim$27M parameters) trained on ImageNet-1K. In all our experiments, we use the AdamW~\cite{AdamW} optimizer with a cosine decay learning-rate scheduler for 80 epochs, preceded by 20 epochs of linear warm-up. In all the experiments, the images are resized to the same fixed resolution (224 × 224). With \method, $\rho$ is set to $50\%$, namely we half the number of neurons in the adapters, at every 100 epochs.

\subsection{Main results}
\label{sec:main_results}
We compare our proposed method \method{} with multiple PETs methods. We remark that all the baselines are obtained with a $C\!=\!5$ cycles training, while \method{} will always have a lower or equal number of training cycles in Tab.~\ref{Multitasktable}. We include the following methods: 

\begin{itemize}[noitemsep, nolistsep]
  \item \textit{Full finetuning}: finetune all parameters of the model.
  
  \item \textit{Att/MLP finetune}: we only tune the Attention/MLP layers and the classification head. 
  
  \item \textit{Linear-probe}: all parameters are frozen except for the task-specific classification layer.
 
  \item \textit{Adapters}~\cite{EfficientNLP}: we add adapters with $\sigma=32$ to have adapters with hidden dimensionality proportional to the input dimension $M_i$. We also include variants where the size of every adapter is fixed over all the layers: $N_i\!=\!47$, and $N_i\!=\!23$. These baselines are considered to emphasize the effect of parameter allocation throughout the layers on the final performance.
 
  \item \textit{BitFit}~\cite{BitFit}: only the biases are finetuned. 
 
  \item \textit{PHM-Adapter}~\cite{PHM}: the weights of the adapters are learned using parameterized hyper-complex multiplication layers (PHM) layers. 
 
  \item  \textit{Compacter}~\cite{Compacter}: adapter weights are learned using shared PHM layers. 
 
 \item  \textit{AdaptFormer}~\cite{AdapterFormer}: introduces Adapters, but only after MLP block with a scaling parameter $s$ applied to the output of the injected modules. 
 
 \item \textit{VPT}~\cite{VPT}: finetuning learnable parameters (i.e. prompts) injected into the embedding space for each layer in ViT.
 \item \imad{\textit{SSF}~\cite{SSF}: aims to adjust the feature activation scaling and shifting its output. 
 }
\item \imad{\textit{Fact-TK}~\cite{FactTK}: a tensorization-decomposition method to store the weight updates into a single 3D tensor. 
 }
 \end{itemize}

\definecolor{large}{RGB}{190, 190, 190}
\definecolor{medium}{RGB}{220, 220, 220}
\definecolor{tiny}{RGB}{240, 240, 240}

\begin{figure}[t]
\begin{center}
\includegraphics[
width=\linewidth]{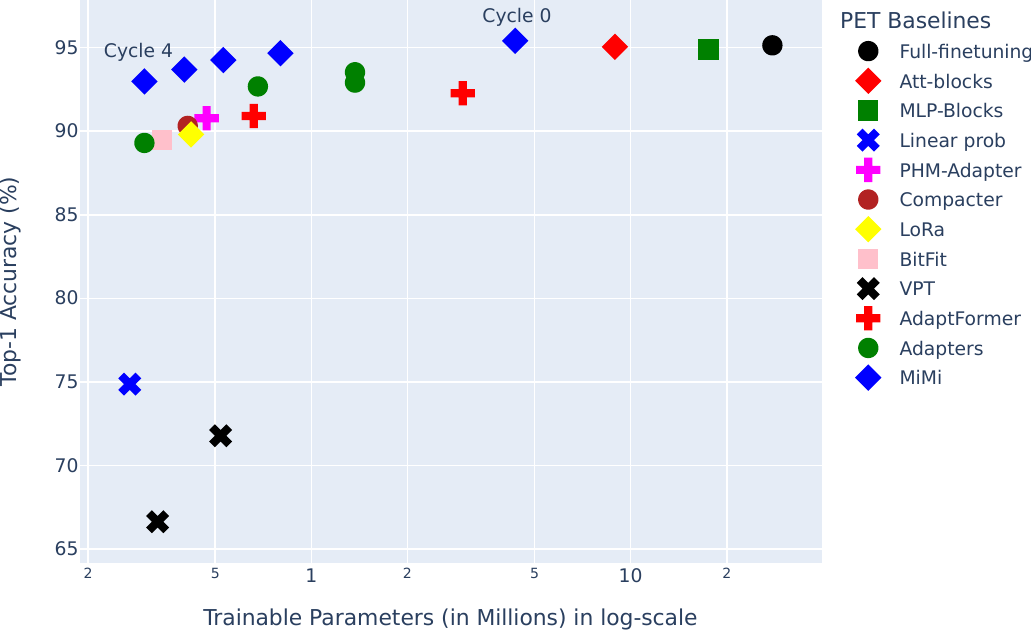}
\end{center}
   \caption{Evaluation of PET baselines mean top-1 accuracy on \textbf{Multi-task benchmark}. We observe \method{} (\textcolor{blue}{\ding{117}}) maintains good performance when reducing the number of parameters, compared to other PET baselines.}
\label{fig:PETs_evaluation}
 \spacefig
\end{figure}

\begin{table}
  \centering
  \resizebox{\linewidth}{!}{
  \begin{tabular}{lcc|ccccc}
    \toprule 
	\textbf{Method} & \textbf{Params}& \textbf{Trained}& \textbf{C100} & \textbf{C10} & \textbf{F.} & \textbf{S.} & \textbf{Mean} \\

    & \textbf{(M)} $\downarrow$&\textbf{(\%)} $\downarrow$&&&&& \\
	
	\midrule 
	{Full finetuning} & \colorbox{large}{27.8} & \colorbox{large}{100} & 88.13 &98.50&97.35&96.59 & 95.14\\ 
	Att-blocks & \colorbox{large}{8.93} & \colorbox{large}{32.14} & 
	88.03 & 98.41 & 97.79 & 95.99 & 95.05\\ 
	MLP-blocks & \colorbox{large}{17.54} & \colorbox{large}{63.12} & 88.44&98.47&96.50&96.14& 94.89
	\\ 

    \textbf{\method}~ (0 cycle)$^\dag$ & \colorbox{large}{4.35} & \colorbox{large}{15.81} & 
	88.27& 98.53& 97.59& 97.28& \textbf{95.41} \\ %
 
	\midrule

    \review{AdaptFormer-64} & \colorbox{medium}{0.66} & \colorbox{medium}{2.38}&
	\review{83.79}& \review{96.93}& \review{90.50}& \review{92.45} &\review{90.91}\\
	
	\review{AdaptFormer-256} & \colorbox{medium}{2.98} & \colorbox{medium}{8.55} &
	\review{84.74} & \review{97.23} & \review{92.13}& \review{94.97}& \review{92.27}\\

    Adapters $N_i\!=\!47$& \colorbox{medium}{1.37} & \colorbox{medium}{4.90} & 
    85.04 & 97.52 & 92.72 & 96.35 & 92.91\\ 
    
    Adapters $N_i\!=\!23$& \colorbox{medium}{0.68} & \colorbox{medium}{2.47} & 
    85.18 & 97.57 & 92.16 & 95.81 & 92.68\\ 
    
    \review{Adapters} $\sigma\!=\!32$& \colorbox{medium}{1.37} & \colorbox{medium}{4.90} & 
	85.59 &97.49 & 94.80 & 96.27& 93.53\\ 
    
    \midrule

Linear prob & \textbf{\colorbox{tiny}{0.27}} & \textbf{\colorbox{tiny}{0.95}} &75.58&91.84&76.80&55.26& 74.87\\ 
 
 PHM-Adapter & \colorbox{tiny}{0.47} &  \colorbox{tiny}{1.72} &
	84.17&96.48&89.18 & 93.32 & 90.78\\  
	
	Compacter & \colorbox{tiny}{0.41} & \colorbox{tiny}{1.44} & 83.95&96.26&88.43&92.67& 90.32 \\ 
	
	
	BitFit & \colorbox{tiny}{0.34} & \colorbox{tiny}{1.22}&
	83.56 &96.14 & 87.85 &90.29&  89.46\\
	
	\review{VPT-deep (10 tokens)} & \colorbox{tiny}{0.33} & \colorbox{tiny}{1.20} &
	\review{67.69}& \review{90.99}& \review{22.77}& \review{85.11}& \review{66.64}\\ 
	
	\review{VPT-deep (100 tokens)} & \colorbox{tiny}{0.52} & \colorbox{tiny}{1.88}&
	\review{72.53}& \review{93.03}& \review{34.88}& \review{86.70} & \review{71.78}\\

    \review{Adapters $N_i\!=\!1$} & \colorbox{tiny}{0.30} & \colorbox{tiny}{1.07}
    & \review{82.60} & \review{96.03} & \review{89.77} & \review{88.80} & \review{89.30} \\

    \imad{SSF} & \colorbox{tiny}{0.28} &  \colorbox{tiny}{0.96} &83.02 & 96.46 & 95.59 & 95.11 & 92.54 \\
    \imad{Fact-TK$_{32}$} & \colorbox{tiny}{0.33} & \colorbox{tiny}{1.18}  & 82.91 & 96.59 & 87.46 & 90.84 & 89.45 \\
    
	\textbf{\method}~ (1 cycle) & \colorbox{tiny}{0.80} & \colorbox{tiny}{2.89} & 
	87.12 &97.98&96.59 & 96.98 & 94.67\\ 

	\textbf{\method}~ (2 cycles) & \colorbox{tiny}{0.53} & \colorbox{tiny}{1.92}&
	86.33&97.49&96.73&96.48 & 94.26\\ 

	\textbf{\method}~ (3 cycles) & \colorbox{tiny}{0.40} & \colorbox{tiny}{1.43} & 
	85.22 & 97.11 & 96.81&95.60& 93.69\\
	
	\textbf{\method}~ (4 cycles) & \colorbox{tiny}{0.30} & \colorbox{tiny}{1.07} & 
	84.07 & 97.11 & 96.81 &93.94& 92.98\\
	\bottomrule
    \end{tabular}
    }
\caption{Results on the \textbf{Multi-task benchmark}. C100, C10, F and S stand for CIFAR100, CIFAR10, Flowers, and SVHN. $^\dag$ is equivalent to \review{Adapters} with $\sigma =8$. Methods are grouped according to the relative number of trainable parameters (\colorbox{tiny}{$\leq 2\%$}, \colorbox{medium}{$\in ]2,10[\%$},\colorbox{large}{$\geq 10\%$}) 
}
\label{Multitasktable}
\spacefig
\end{table}

\noindent\textbf{Discussion.} Fig.~\ref{fig:PETs_evaluation} visualizes the average accuracy versus the number of trainable parameters achieved for the Multi-task benchmark, while Table~\ref{Multitasktable} reports the number of trained parameters and the average accuracy across datasets in the MultiTask benchmark. The detailed Tables for DomainNet and VTAB benchmarks are in the supplementary material. For all the benchmarks, the number of trained parameters is reported in millions, and the average top-1 accuracy on the datasets is reported in the rightest column.

We observe that \emph{full finetuning} achieves commendable performance, albeit demanding an extensive parameter-tuning for each dataset. In comparison, finetuning solely the \emph{attention/MLP} layer proves remarkably effective among the vanilla finetuning baselines. However, this approach still necessitates a substantial number of task-specific parameters, unlike other PET approaches. Notably, the underwhelming performance of \emph{linear probing} emphasizes the significance of altering the feature representations within the model when adapting to new tasks.

Notably, both \emph{PHM} and \emph{Compacter} demonstrate their effectiveness by achieving impressive performance while adjusting less than 2\% of the parameters. Unlike NLP tasks where PETs have shown success with a small number of trainable parameters~\cite{EfficientNLP}, visual tasks do not attain \emph{full finetuning} performance with such limited parameter adjustments. Additionally, the subpar performance of \emph{VPT} indicates that injecting tokens into the embedding space offers minimal benefit when the pre-training dataset differs from the downstream task. Remarkably, all PET methods consistently maintain similar performance rankings across all tasks, suggesting that the optimal adaptation strategy is independent on the specific downstream task.

\emph{Adapters} achieve impressive results with a slightly higher number of trainable parameters (1.37M, 4.90\% of the total) for $\sigma\!=\!32$. Remarkably, \emph{Adapters} outperform \emph{AdaptFormer}~\cite{AdapterFormer} while utilizing fewer parameters (92.91\% with 1.37M parameters compared to 92.27\% with 2.98M parameters). This outcome highlights the superiority of adapting representations after both MSA and MLP blocks, as demonstrated in the architecture of \emph{Adapters} (Fig.~\ref{fig:main}), over solely acting on the MLP block, as in done \emph{AdaptFormer}.

\begin{figure}[t]
\begin{center}
\includegraphics[trim={15 27 45 70}, clip, width=\linewidth]{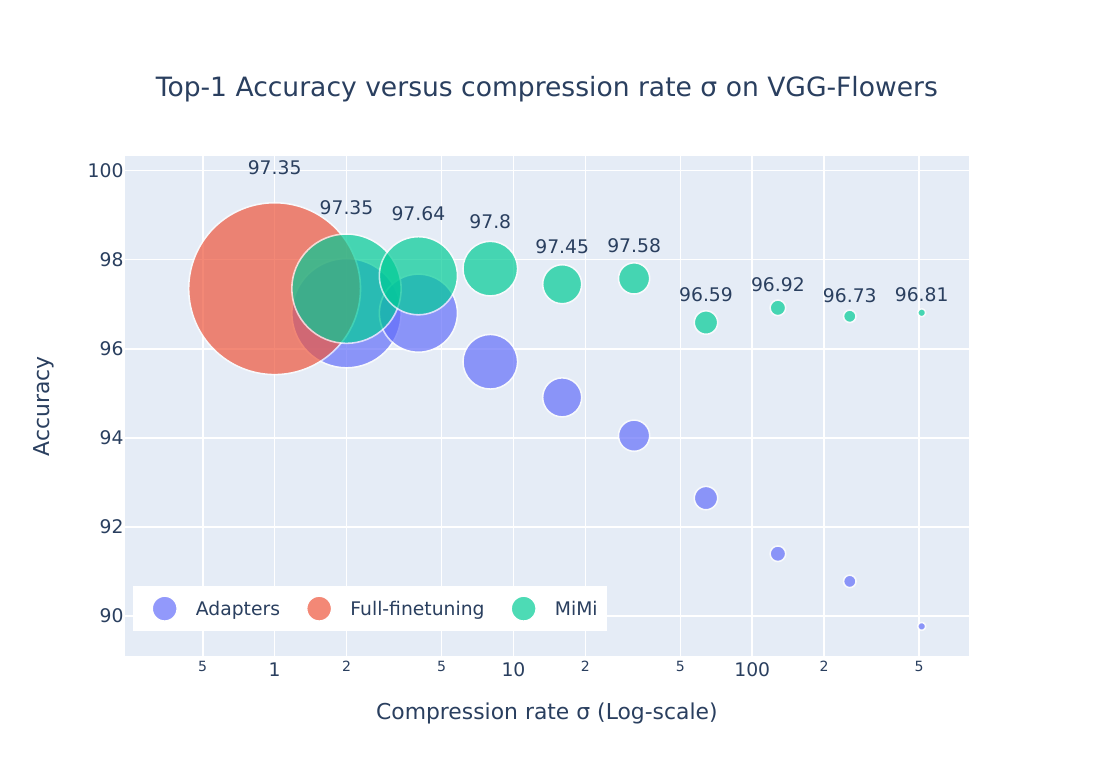}
\end{center}
    \vspace{-4mm}
   \caption{Comparison of top-1 accuracy versus compression rate $\sigma$ on VGG-Flowers. \imad{All MiMi results originate from the same MiMi run. Adapters are trained for the exact same number of epochs as their MiMi counterparts}. The size of blob markers represents the number of trainable parameters.}
\label{fig:TINA_performance}
 \spacefig
\end{figure}

We observe that \emph{\method} significantly reduces the number of parameters by 4 times the initial size (0.40M, $1.43\%$) while outperforming all PET methods in the Multi-task benchmark. In particular, \method{} outperforms \emph{adapters-$n_{i}\!=\!47$} despite having fewer parameters, demonstrating that our iterative training procedure improves the parameter efficiency. To further emphasize the performance gap between the two approaches, we introduce Fig.~\ref{fig:TINA_performance} illustrating the performance as a function of the number of trainable parameters for VGG-Flowers (for CIFAR-100 dataset in supplementary). We observe the significant performance gap between vanilla adapters compared to adapters trained with the \method{} approach.

Furthermore, \method{} outperforms methods with similar trained parameters, in all the compression ranges. In particular, in the most challenging one (with 0.30M parameters), \method{} outperforms the closest approach, BitFit, which trains 0.34M parameters, showing a gain in average accuracy larger than $3\%$ and $2\%$, for Multi-Task and DomainNet benchmarks, respectively.

Upon comparing \emph{adapter} with uniform and proportional parameter distribution ($N_i$ vs $\sigma$), results are in favor of allocating parameters proportionally to the layer dimension. Notably, adapters with $\sigma\!=\!32$ outperform adapters with $N_{i}\!=\!47~\forall i$ in both the Multi-task (93.53\% vs 92.91\%) and DomainNet (70.65\% vs 69.39\%) benchmarks. This observation suggests that the task-specific nature of the last layers, with higher dimensionality, necessitates greater adaptation. Furthermore, we demonstrate that reducing the size of adapters ($N_{i}\!=\!23$) negatively affects performance, with a marginal drop in Multi-task (0.23\%) but a more consistent decrease in DomainNet (1.01\%). These findings underscore \textit{the unsatisfactory performance obtained from training adapters in a vanilla fashion and serve as motivation for our specific training procedure}.\\
\noindent\textbf{\method{} versus Vanilla training.} 
Looking at the Multi-task benchmark (Fig. \ref{fig:PETs_evaluation}, Table~\ref{Multitasktable}), we observe that \emph{\method} significantly reduces the number of parameters by 4$\times$ (0.40M, $1.43\%$) while outperforming all PET methods in the Multi-task benchmark. In particular, 
\emph{\method{}} outperforms \emph{adapters-$N_{i}=47$} despite having fewer parameters, demonstrating that our iterative training procedure improves the parameter efficiency. To further emphasize the performance gap between the two approaches, we introduce Fig.~\ref{fig:TINA_performance}, we observe \textit{the significant performance gap between vanilla adapters compared to adapters trained with \method{} approach}.

\subsection{Ablation study}
\label{sec:ourscore}

\paragraph{Importance score for \method{}.} Next, we move on to our design choice of dimensionality reduction inside adapters throughout the adaptation cycles. We report the contribution of various components of \method{} with different setups. 
\begin{itemize}[noitemsep,nolistsep]
    \item \emph{Vanilla Adapters}: corresponds to injecting adapters with a compression rate $\sigma$.
    
    \item \textit{Random}: we select randomly a percentage 
    of neurons for each adapter to be removed. 
    
    \item \textit{Gradient-Based $\normlone(\nabla)$}: We determine the neurons to be removed based on the $\normlone$ norm of the gradients. 
    
    \item \textit{Local neuron selection}: We uniformly choose a percentage of neurons to be removed, independently applied to both down-sampling and up-sampling layers.

    
     \item \textit{Global neuron selection}: The number of neurons to be removed per adapter is determined using \eqref{eq:impscore} given $\rho$, considering the scaling factor if applicable. Additionally, we assess our scoring function without the inclusion of the scaling factor $n_i+m_i$. This modified version of our score is referred to as $\mathcal{I}_{0}$.      
     
     \item \textit{\method{}}: our method as in Alg.~\ref{alg:ourmethodsimplified}.
\end{itemize}

\noindent To compare the different methods we proceed as follows. When using an iterative method, we always start from the baseline model where $\sigma\!=\!32$. When using a non-iterative method: we start with adapters of $\sigma_{0}\!=\!\sigma_{target}/(1-\rho)$ and prune once only after the first cycle. Training continues for $C\!-\!1$ cycles to guarantee fair comparison with iterative methods. Results are reported in Table \ref{OursAblationStudy}.

\begin{table}
\centering
      \centering
      \renewcommand{\arraystretch}{1.3}
      \resizebox{\linewidth}{!}{
      \begin{tabular}{ccccc ccccc}
      \toprule
        \multirow{2}{*}{\textbf{Method}}&\multirow{2}{*}{\textbf{Selection}} &
        \multirow{2}{*}{\textbf{Score}} & 
        \multirow{2}{*}{\textbf{Iter.}}& \multirow{2}{*}{\textbf{Scale}}&\multicolumn{5}{c}{$\boldsymbol{\sigma}$} \\ 
      & &&&&\textbf{$32$} & \textbf{$64$} & \textbf{$128$} &\textbf{$256$}&\textbf{$512$}  \\
        \midrule
        
        \footnotesize{\textbf{Vanilla}} &- &-&-&-&94.80 & 90.12& 89.42& 88.85 & 86.03 \\
        \midrule
        
      \multirow{11}{*}{\footnotesize{\textbf{Baseline}}}& 
      \review{Random} &$-$&\checkmark&&- & 95.43 & 95.80 & 95.11 & 95.12 \\
      
      &Local (DW) &$\normlone(w)$&&&- & 95.46 & 95.06 & 94.28 &93.79 \\
     &\review{Local} &$\normlone(\nabla)$&\checkmark&&- & 95.46 & 96.17 & 96.11 & 96.42 \\
     &Local & SA~\cite{SparseAdapters} &&&- & 96.10 & 95.57 & 96.15 &96.23 \\
     &Local & SA~\cite{SparseAdapters} &\checkmark&& - &96.41 & 96.65 & 96.72 & 96.72 \\
     &  Local& \imad{GraSP~\cite{GraSP}} & & & & 90.62 & 89.71 & 87.22 & 86.66 \\ 
     &  Local & \imad{SNIP~\cite{SNIP}}  & & & & 93.53 & 92.39 & 91.36 & 90.62 \\ 
     
    &\review{Global} & \review{$\normlone(a)$}&&\review{\checkmark}&\review{-} &\review{96.10}& \review{94.28}& \review{93.80}& \review{93.25}\\
     &\review{Global} & \review{$\normlone(a)$}&\review{\checkmark}&\review{\checkmark}&\review{-} & \review{96.13}& \review{95.15}& \review{95.77}& \review{95.72}\\
     &Global & $\mathcal{I}_{0}$&&&- & 94.88 & 95.28 & 95.66&95.45 \\
     &Global & $\mathcal{I}$&&\checkmark&- & 96.10 & 95.82 & 96.34&96.50\\
     \midrule
     \textbf{\footnotesize{\method{}}}& Global &  $\mathcal{I}$&\checkmark&\checkmark& - & \textbf{96.59} & \textbf{96.92} & \textbf{96.73}& \textbf{96.81}\\
    \bottomrule
        \end{tabular}}
        \caption{Performance analysis 
    for neuron selection on VGG-Flowers. $\normlone(w)$ and $\normlone(a)$ denote the magnitude pruning of the parameters and the activations respectively. Local (DW) represents local pruning applied to down-sampling layers only.}
    \label{OursAblationStudy}
     \spacefig
\end{table}
\noindent\textbf{Discussion.} Table~\ref{OursAblationStudy} summarizes the performance of local and global neuron selection for adapters.
Firstly, we observe that reducing the number of parameters in vanilla adapters (higher values of $\sigma$ in Fig.~\ref{fig:TINA_performance}) leads to a drop in performance. Additionally, we find that using the magnitude of parameters instead of activations is advantageous. Activation-based scoring methods rely on input images, requiring batch or dataset-level statistics, which are computationally less efficient.

Secondly, global neuron selection proves to be superior to local neuron selection. The former method focuses on finetuning adapters in specific layers while completely removing other adapters, while the latter removes the same amount of neurons from each adapter's layer. Moreover, \method{} surpasses SA by employing structured pruning (neuron removal) instead of unstructured pruning (weight removal), to reduce the size of adapters. Additionally, \method{} incorporates a look-ahead strategy that accounts for the impact of up-sampling layers, ensuring consistently high performance. Notably, with MiMi, adapter size can be reduced for efficient computation, unlike SA.

In the final cycles, \method{} identifies crucial adapters for adaptation, prioritizing their finetuning. This approach improves model performance by adjusting specific latent representations tailored to the downstream task while using fewer parameters. \imad{ \method{} consistently outperforms both GraSP and SNIP across all $\sigma$ values due to its iterative pruning approach. Pruning at initialization as done in SNIP/GraSP, before the adapters have been trained are less effective. Since the initialization is random, they are missing out on retaining potentially important weights.}

Table~\ref{OursAblationStudy} reveals that \method{} achieves superior performance compared to vanilla adapter training on the VGG-Flowers dataset, with a performance gap of $6.14\%$, when using $\sigma=64$ (regardless of local/global neuron selection). Notably, this performance gap increases as we reduce the adapter size to $\sigma=256,512$. Furthermore, when comparing to a vanilla $\normlone$ importance scoring, we observe the benefits of considering both down-sampling and up-sampling parameters for the adapters. This approach consistently improves performance across compression rates ranging from $0.5\%$ to over $3\%$. Notably, the performance gap becomes more prominent at higher compression rates.

Finally, scaling the importance score according to \eqref{eq:impscore} enhances the performance of the \emph{Global} method by approximately $1\%$ across all $\sigma$ values.

\begin{figure*}[t]
\centering
\includegraphics[width=\textwidth]{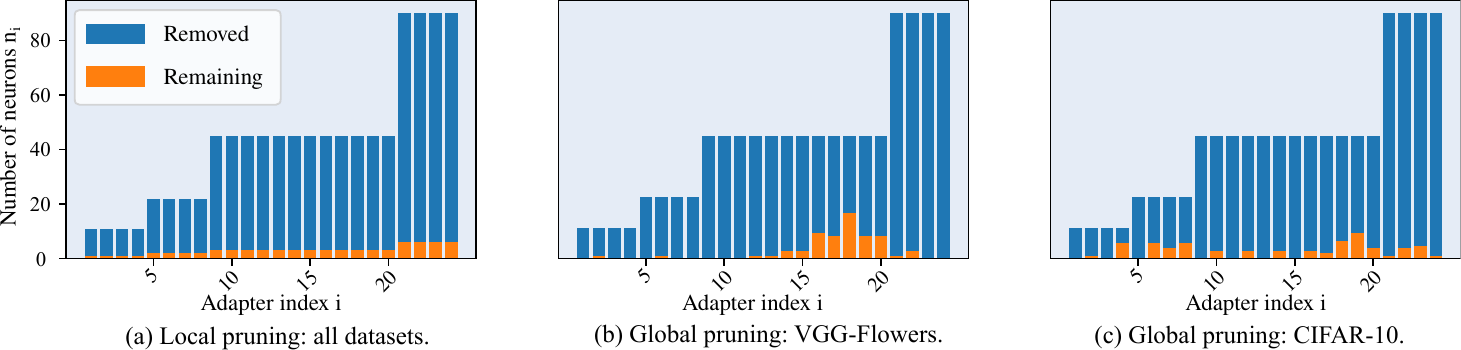}
\caption{Layer-wise analysis of adapter's neurons distribution at $4^{th}$ cycle. Bar plots represent the number of neurons $n_i$ at each adapter $i$ for VGG-Flowers and CIFAR-10, respectively. Global neuron selection leads to different neuron distributions depending on the dataset. Compared to VGG-Flowers, fewer adapters are completely removed on CIFAR-10.}
\label{fig:globa_vs_local}
\end{figure*}

\noindent\textbf{Parameter allocation analysis with \method{}.}
In Fig.~\ref{fig:globa_vs_local}, we visualize the distribution of removed and remaining neurons achieved through the application of \method{} on VGG-Flowers and CIFAR-10. Notably, this illustration highlights the contrasting outcomes between local neuron selection, which uniformly removes neurons from adapters, and the utilization of global neuron selection. Remarkably, we observe that the latter approach completely removes certain adapters from the model (evident in layers $4, 5, 7,$ and $8$ of VGG-Flowers), while redistributing a significant number of parameters to other adapters.

Moreover, \emph{global neuron selection} exhibits distinct adaptations for each dataset, as evidenced in Fig.~\ref{fig:globa_vs_local}. Notably, the distribution of removed neurons varies between CIFAR-10 and VGG-Flowers. In the case of CIFAR-10, fewer adapters are completely removed compared to VGG-Flowers. Conversely, for VGG-Flowers, only adapters at later stages are retained, suggesting that early layer representations are well-suited for this particular dataset. However, for CIFAR-10, the remaining adapters are dispersed throughout all layers of the ViT model, indicating that almost all the layers' representations need to be finetuned. These observations highlight the adaptability and dataset-specific optimizations achieved through \emph{global neuron selection}. To provide a more comprehensive analysis, we also present normalized plots in supplementary material. 

\noindent\textbf{ViT variants with \method{}.}
We evaluate the performance of \method{} on different ViT backbones, including Swin-S/L ($\sim$50M/$\sim$197M  parameters), ViT~\cite{ViT} ($\sim$86M parameters), and CVT~\cite{CvT} ($\sim$20M parameters).

For three training cycles, we compare the three baselines: \emph{finetuning}, \emph{adapters}, and \emph{\method{}}. Table~\ref{BackboneViTs} presents the best scores achieved in the final cycle. Remarkably, \method{} achieves comparable performance to full model finetuning, with a margin of 1.2\%, 1.2\%, and 1.4\% for ViT-B/16, Swin-S, and CvT, respectively. This is accomplished by finetuning less than 1.5\% of the parameters, including the head classifier. \method{} surpasses vanilla adapters' performance with four times fewer parameters across all ViT backbones (ViT, Swin-T, and CvT). These experiments demonstrate the generalizability of \method{} to various ViT backbones.

\begin{table}[t]
  \centering
  \renewcommand{\arraystretch}{1.1}
  \resizebox{\linewidth}{!}{
  \begin{tabular}{lcccccccc}
    \toprule
	& \textbf{Method} & \# \textbf{Params}& \textbf{Trained}& \textbf{C100} & \textbf{C10} & \textbf{V} & \textbf{S} & \textbf{Mean}\\

     && \textbf{(M)} $\downarrow$&\textbf{(\%)} $\downarrow$&&&&&$\uparrow$\\
	\midrule
	\multirow{4}*{\rotatebox{90}{ViT-B-16}}
	& Finetune & 85.90 & 100 & 91.22 &	99.01 &	99.32 & 97.68 & \textbf{96.81}\\
	\cmidrule{2-9}
	& Adapters & 0.96 & 0.89 & 89.39 & 98.02 &	97.69 &	94.17 &	94.82 \\
	& \method{} & 0.62 & 0.54 & 89.86 & 98.09 & 98.75 &	94.94 & 95.41 \\
        & \method{} & \textbf{0.37} & \textbf{0.32} & 89.84 & 98.17 & 98.85 &	95.32 & \underline{95.55} \\
	\midrule
	\multirow{4}*{\rotatebox{90}{Swin-S}}
	& Finetune & 48.80 & 100 & 90.12&	98.88&	98.37 & 98.16 & \textbf{96.38}\\
	\cmidrule{2-9}
	& Adapters & 0.41 & 4.88 & 89.05 & 98.48 &	94.60 &	97.25 &	94.84 \\
	& \method{} & 0.23 & 2.75 & 88.86 & 98.53 & 96.16 &	97.22 & 95.19 \\
    & \method{} & \textbf{0.11} & \textbf{1.32} & 88.62 & 98.50 & 96.68 &	96.94 & \underline{95.18} \\
	\midrule

\multirow{4}*{\rotatebox{90}{\imad{Swin-L}}}
	& Finetune & 197M & 100\% & 95.12 &99.34 & 99.67 & 98.22 & 98.08\\
	\cmidrule{2-9}
	& Adapters & 20.1M & 10.2\% & 94.31 &99.46 & 99.76 & 97.98 & \underline{97.88} \\
	& \method{} & 10.9M  & 5.53\% & 94.78 & 99.44 & 99.51 & 99.77 & \textbf{98.38}\\
        & \method{} & 6M  & 3.04\% & 92.92 & 99.30 & 99.74 & 97.96 & 97.48 \\
	\midrule
 
	\multirow{4}*{\rotatebox{90}{CvT}}
	& Finetune & 19.65 & 100& 90.01& 98.68 &	97.98 &	98.09 & \textbf{96.19}\\
	\cmidrule{2-9}
	& Adapters & 0.78 & 4.00 & 86.68 & 97.91 & 88.93 & 96.96 & 92.62\\
	& \method{} & 0.47 & 2.40 & 86.47 & 97.98 & 93.28 &	97.17 & \underline{93.73}\\
    & \method{} & \textbf{0.28} & \textbf{1.44} & 85.87 & 97.77 & 94.31 &	96.67 & 93.66 \\
	\bottomrule
\end{tabular}
}
\caption{Performance of our method using different ViT backbones on the Multi-task benchmark. The highest score is in bold and the second best score is underlined. C100, C10, F, and S stand for CIFAR100, CIFAR10, Flowers, and SVHN datasets.}
\label{BackboneViTs}
\spacefig
\end{table}
\noindent\textbf{Evaluating Inference Cost/Storage Footprint.}
In this section, we conduct a comprehensive analysis of the GFLOPS at inference time and storage footprint for PETs methods in the context of multi-task learning. Table~\ref{FootPrintTable} presents the findings, including the number of trainable parameters and the storage requirement in MegaBytes (MB) for saving the Swin-T model after finetuning per task $T$.

Storing a complete ViT for each task can impose significant burdens on storage space and computational resources. With ViTs containing millions of parameters, these storage requirements quickly accumulate in multi-task settings. However, by storing only a subset of the model's parameters, both storage costs and computational resources for training can be significantly reduced.

\begin{table}[t]
  \centering
  \renewcommand{\arraystretch}{1.1}
  \resizebox{\linewidth}{!}{
    \begin{tabular}{lcccc}
    \toprule 
	\textbf{Method} & \# \textbf{Params (M)$\downarrow$}& \textbf{Storage (MB)$\downarrow$} & \textbf{GFLOPS $\downarrow$}&\textbf{Accuracy(\%) $\uparrow$}\\
	\midrule 
        Full-finetuning & 27.8 &  111 & 8.72 & 97.35\\  
        Att-block & 8.93  &  34.7 & 8.72 & 97.79 \\   
        MLP-blocks & 17.54  &  73.4 & 8.72 & 96.50 \\  
        Full-model W/ Adapter($\sigma =32$) & 1.37 & 115.3 & 9.06 & 96.27\\
        \midrule
	BitFit &  0.34 &  0.34   & 8.72 & 87.85\\ 
	VPT-Deep (100 tokens) &  0.32 & 160.1 & 18.40 & 34.88\\
	AdaptFormer-64 & 0.84 & 1.63 & 9.08 & 90.50 \\ 
        SSF & 0.28 & 0.96  &  8.72 &  95.59\\
        Fact-TK$_{32}$ & 0.33 & 1.18  &  10.6 &  87.46\\
        
        Adapters($\sigma =32$) & 1.37 & 4.30 & 9.06 &96.27 \\
        Adapters($n_i=47$) & 1.37 & 4.40 & 9.26 &92.72 \\
        Adapters($n_i=1$) & 0.30 & 0.47 & 8.74 &89.77 \\
        \midrule
        Linear-prob & 0.27  &  0.31 & 8.72 &76.80 \\  
        \textbf{\method{} (3 cycles)} & 0.40 & 0.63 & 8.92 &96.81 \\
	\textbf{\method{} (4 cycles)} & 0.30 & 0.47 & 8.82 &96.81 \\
	\bottomrule
    \end{tabular}
    }
    \caption{Comparison with different PET methods on VGG-Flowers with respect to inference cost (GFLOPs).}
    \label{FootPrintTable}
    \spacefig
\end{table}

Larger model components, like attention and MLP blocks, demand significant storage due to their extensive trainable parameters. On the other hand, finetuning only the head (\textit{Linear-prob}) is lightweight but comes at the cost of performance. Notably, \method{} achieves a +6\% performance enhancement compared to Adapters $n_{i}\!=\!1$ while maintaining similar storage needs. 
However, \method{} offers both robust performance and reduced memory demands, positioning it as a superior alternative.

In terms of GFLOPS during inference, full fine-tuning, and similar variants, such as Att-block and MLP-blocks, achieve the lowest GFLOP values at 8.72. However, they come at the expense of a high memory footprint. On the other hand, VPT-Deep (100 tokens) stands out with the highest GFLOPS at 18.40, thanks to an increase in the embedding space for each layer to 100 tokens. This emphasizes that fewer parameters do not necessarily guarantee computational efficiency. \method{} in its 3-cycle and 4-cycle variants, achieves GFLOPS values of 8.92 and 8.82, respectively. This efficiency is attributed to our method's ability to completely remove some adapters, effectively reducing the computational cost during inference.

\section{Conclusion}
In this work, we propose \method{}, a training algorithm to learn small adapters for the problem of ViT efficient finetuning.  Rather than directly training adapters with few parameters, we propose to start with large adapters, and then iteratively select the more important neurons in every adapter. Our training procedure estimates the hidden dimension for each adapter, reducing the number of trainable parameters and even removing adapters if unnecessary. We empirically demonstrate the greater performance of \method{} to adapters and show that our method achieves excellent performance with low numbers of trainable parameters. Our ablation study validates the positive impact of our novel importance score to estimate the hidden dimension of each adapter. 
\newline
\noindent\textbf{Acknowledgements. }This paper has been supported by the
French National Research Agency (ANR) in the framework of
its JCJC. Furthermore, this research was partially funded by Hi!PARIS Center on Data Analytics and Artificial Intelligence.

\newpage

{\small
\bibliographystyle{ieee_fullname}
\bibliography{egbib}
}

\clearpage
\setcounter{page}{1}
\section*{Supplementary Material}

\maketitle


In this supplementary material, we provide \imad{(1) a visual representation of our method MiMi for clarity (2) additional experimental results to further analyze the proposed \method{} approach (Multi-task benchmark, and VTAB benchmark). (3) justify in more detail our choice for the design of the importance score (4) the effect of the parameters allocation in Adapters for ViTs (5) provide details regarding the datasets used in our experiments.}

\section{Illustration of \method{} Design}
In our work, we augment a pre-existing ViT model with an 'adapter' module. For each adapter, the input is denoted as $\boldsymbol{h}_i$ with dimension $M_i$. The adapter undergoes two primary transformations.

Firstly, it maps $\boldsymbol{h}_i$ to $\boldsymbol{z}_i \in \R^{N_i}$ through a fully-connected layer characterized by a parameter matrix \( \boldsymbol{W}_i^{\text{down}} \). A non-linear activation function \( \phi(\cdot) \) is also applied during this step. Subsequently, $\boldsymbol{z}_i$ is transformed back to an output $\boldsymbol{r}_i\in \R^{M_{i}}$ by another fully-connected layer, parameterized by \( \boldsymbol{W}_i^{\text{up}} \).

A special feature of the adapter is its residual skip-connection. If \( \boldsymbol{r}_i \) is near zero, the adapter essentially acts as an identity function, making minimal changes to the input.

\begin{figure}[h]
  \centering
  \includegraphics[width=0.8\linewidth]{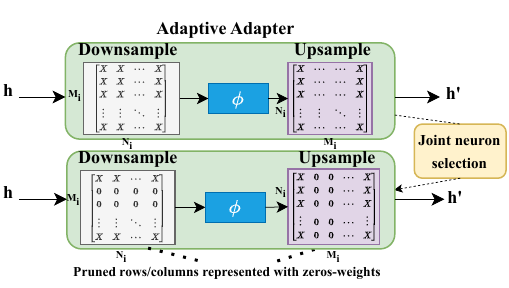}
   \caption{Illustration of the design of adapter after using MiMi.}
   \label{fig:illustration_mimi}
   \vspace{-4mm}
\end{figure}

Our design of \method{} is influenced by an observation: if an entire row in \( \boldsymbol{W}_{i}^{\text{down}} \) and an entire column in \( \boldsymbol{W}_{i}^{\text{up}} \) are zero, the adapter behaves as though its complexity is reduced, effectively acting as if it has a smaller dimension $M_i$, as illustrated in Fig.\ref{fig:illustration_mimi}.

\section{\method{} versus Vanilla training on CIFAR-100} 
\begin{figure}[b]
    \centering
    \includegraphics[trim={15 27 45 70}, clip, width=\linewidth]{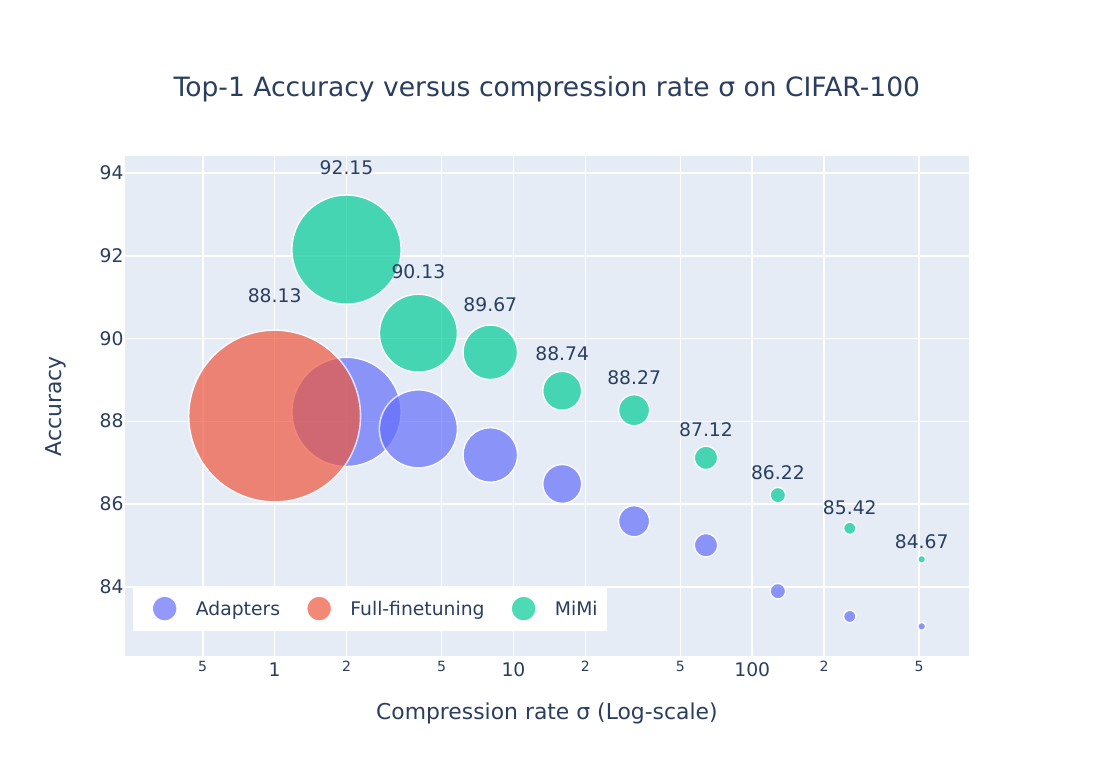}
    \caption{\review{Comparison of top-1 accuracy of vanilla adapters, and \method{} with respect to compression rate $\sigma$ on CIFAR-100 dataset. All MiMi results originate from the same MiMi
    run. Adapters are trained for the exact same number of epochs as their MiMi counterparts. The size of blob markers represents the number of trainable parameters. We notice that at $\sigma=2, 4, 8$, \method{} outperforms full finetuning.}}
\label{fig:TINA_performance2}
\end{figure}

Looking at Fig.~\ref{fig:TINA_performance2}), we observe \textit{the significant performance gap between vanilla adapters compared to adapters trained with \method{} approach}. Our method outperforms vanilla adapters with a more than 10\% accuracy gap at small sizes.

\section{Local versus Global Neurons Selection}

\begin{table*}[t]
    \caption{A comparative performance analysis of local and global neuron selection on VGG-Flowers.}
      \begin{small}
      \label{OursAblationStudy2}
      \centering
      \renewcommand{\arraystretch}{1.2}
      \resizebox{\textwidth}{!}{
      \begin{tabular}{lcccc ccccccccc}
      \toprule
        \multirow{2}{*}{\textbf{Method}}&\multirow{2}{*}{\textbf{Neurons Removal}} & \multirow{2}{*}{\textbf{Selection}} &  \multirow{2}{*}{\textbf{Iterative}}& \multirow{2}{*}{\textbf{Scaling}} & \multicolumn{8}{c}{$\boldsymbol{\sigma}$} \\
        &&&&&\textbf{$32$} & \textbf{$64$} & \textbf{$128$} &\textbf{$256$}&\textbf{$512$}
        &\textbf{$1024$}&\textbf{$2048$} & \textbf{$4096$}  \\
        \midrule
        \multirow{6}*{\rotatebox{90}{}}
        \review{\textbf{Vanilla adapters}}  &  && & &94.80 & 90.12& 89.42& 88.85 & 86.03 & 86.09 & 85.14 & 85.14\\
        \midrule
        \multirow{4}{*}{\textbf{Baselines}}& Local &\checkmark&&&- & 96.10 & 95.57 & 96.15 &96.23 & 96.76 & 95.83 & 95.83\\
        & Local & \checkmark &\checkmark&& - &96.41 & 96.65 & 96.72 & 96.72 & \textbf{96.81} & \textbf{96.83} & 94.54\\
        & Global & \checkmark &&&- & 94.88 & 95.28 & 95.66&95.45 & 95.56 & 96.03 & 96.03 \\
        & Global& \checkmark &&\checkmark&- & 96.10 & 95.82 & 96.34&96.50 & 96.15 & 96.03 & 96.03 \\
        \midrule
        \textbf{\method{}} & Global &\checkmark &\checkmark&\checkmark& - & \textbf{96.59} & \textbf{96.92} & \textbf{96.73}& \textbf{96.81} &96.55 & 96.47 & \textbf{96.17}\\
         \bottomrule
        \end{tabular}}
         \end{small}
\end{table*}

        \begin{table*}[t]
      \begin{small}
      \centering
        \caption{A comparative performance analysis of local and global neuron selection on CIFAR-10 and CIFAR-100.}
        \label{OursAblationStudy2_cifar}
      \renewcommand{\arraystretch}{1.2}
              
        \resizebox{\textwidth}{!}{\begin{tabular}{lcccc ccccccccc}
      \toprule
        \multirow{2}{*}{\textbf{Method}}&\multirow{2}{*}{\textbf{Neurons Removal}} & \multirow{2}{*}{\textbf{Selection}} &  \multirow{2}{*}{\textbf{Iterative}}& \multirow{2}{*}{\textbf{Scaling}} & \multicolumn{8}{c}{$\boldsymbol{\sigma}$} \\
        &&&&&\textbf{$32$} & \textbf{$64$} & \textbf{$128$} &\textbf{$256$}&\textbf{$512$}
        &\textbf{$1024$}&\textbf{$2048$} & \textbf{$4096$}  \\
        \midrule
        &&&&& \textbf{CIFAR-10} \\
        \midrule
        \review{\textbf{Vanilla adapters}}& -   &&& & 97.89 & 97.39& 97.06& 96.60 & 96.53 & 96.27 & 82.06 & 82.06\\
        Baseline & Local & \checkmark &\checkmark&& - & \textbf{98.06} & 97.92 & \textbf{97.64} & 97.11 & \textbf{96.91} & \textbf{96.44} & 93.49\\
        & Random & \checkmark &\checkmark&& - & 97.79 & 97.68 & 97.42 & 96.98 & 96.56 & 96.15 & 93.84\\
        & Gradient & \checkmark &\checkmark&& - & 97.76 & 97.73 & 97.56 & 97.14 & 96.71 & 96.29 & 93.92\\
         \textbf{\method{}} &Global& \checkmark &\checkmark&\checkmark& - & 97.98 & 97.92 & 97.49 & \textbf{97.15} & 96.71 & 96.04 & \textbf{95.57}\\
        \hline
        
        \multirow{6}*{\rotatebox{90}{}}
        &&&&& \textbf{CIFAR-100} \\
        \midrule
        \review{\textbf{Vanilla adapters}} & -  &&& & 86.22 & 85.02& 84.33& 83.54 & 82.26 & 82.19 & 83.17 & 82.19\\
        \textbf{Baseline} & Local & \checkmark &\checkmark&& - & 86.88 & 86.15 &85.30 & 84.06 & 83.71&\textbf{83.35} & 78.44\\
        & Random & \checkmark &\checkmark&& - & 85.97 & 85.71 & 84.82 & 84.23 & 83.72 & 83.08 & 78.24\\
        & Gradient & \checkmark &\checkmark&& - & 86.11 & 85.67 & 85.16 & 84.57 & 84.16 & 83.35 & 78.35 \\
         \textbf{\method{}} &Global& \checkmark &\checkmark&\checkmark& - & \textbf{87.12}&\textbf{86.22}&\textbf{85.42}&\textbf{84.67}&\textbf{83.25}&82.67&\textbf{82.10}\\
        \bottomrule
        \end{tabular}}
        \end{small}
\end{table*}

Here below, we provide extra experiments for adapters with different sizes for VGG-Flowers, CIFAR-10, and CIFAR-100.

Tables~\ref{OursAblationStudy2} and~\ref{OursAblationStudy2_cifar} report the performance of \method{} algorithm with respect to vanilla training -\emph{Baseline}- on VGG-Flowers, CIFAR-10 and CIFAR-100. \method{} outperforms vanilla adapters on all the adapters with a significant performance gap. Interestingly, this gap in performance expands, as we compare smaller adapters $\sigma=256,...,4096$ (higher compression ratio).

Furthermore, these results emphasize the usefulness of each component of the importance score for \method{}. We notice that applying global neuron selection, normalization, and iterative training outperforms local neuron selection on almost all adapter sizes for VGG-Flowers (Table~\ref{OursAblationStudy2}) and CIFAR-100 (Table~\ref{OursAblationStudy2_cifar}). This indicates that each component of the importance score of \method{} is important to boost performance and reduce the parameters.

\section{Vanilla versus \method{} Training for Adapters}

\begin{figure}[t]
    \centering
    \includegraphics[trim={25 0 0 10}, clip,width=\linewidth]{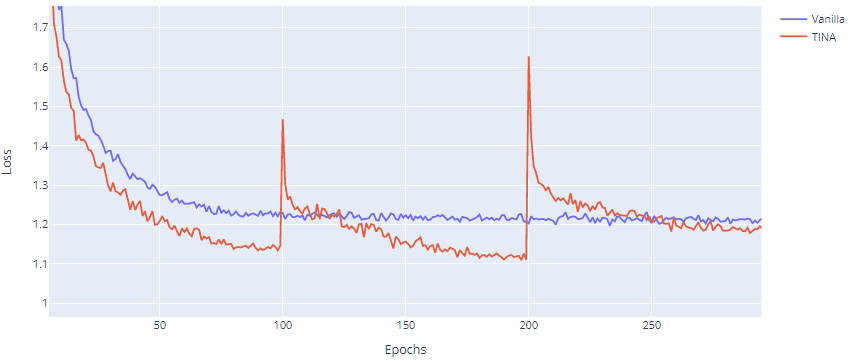}
    \caption{Training loss curves of finetuning adapters with vanilla, and \method{} training on CIFAR-100 dataset.}
\label{fig:loss1}
\end{figure}

\begin{figure}[t]
    \centering
    \includegraphics[trim={25 0 0 10}, clip,width=\linewidth]{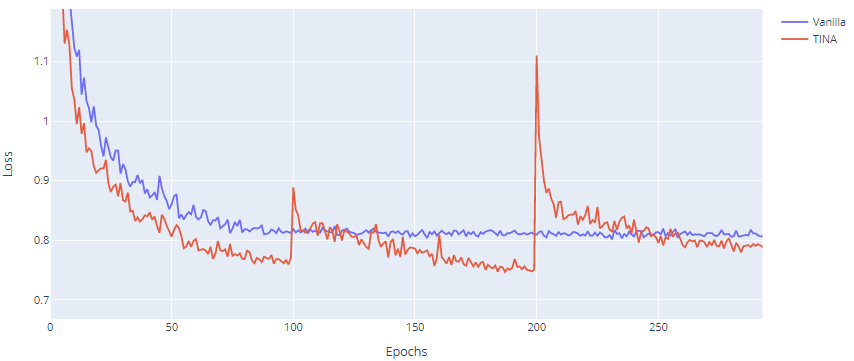}
    \caption{Training loss curves of finetuning adapters with vanilla, and \method{} training on SVHN dataset.}
\label{fig:loss2}
\end{figure}

To validate the greater optimization performance of \method{}, in Figs.~\ref{fig:loss1}, \ref{fig:loss2}, we show the training loss curves of vanilla, and \method{} training of adapters for CIFAR-100, and SVHN, respectively. At the end of the training, the two models (\ie~ vanilla training and \method{}) have similar numbers of training parameters.

We notice that the loss of the training using \method{} algorithm is much smoother than vanilla training resulting in adapters that generalize well on the downstream task as previously shown in Fig. 4. Furthermore, we notice spikes in the training loss of \method{}, due to the removal of neurons after each cycle. Eventually, sequential training and neuron selection is a successful strategy to find small networks while maintaining good performance, since directly training small networks does not provide similar results ~\cite{frankle2018lottery}.

\section{Impact of the parameter allocation}
\label{NoTStages} 

In this ablation study, we validate the idea that every layer of a ViT needs to be adapted differently in order to learn a new task. The Swin-B vision transformer model that we use in all our experiments consists of 4 stages. Therefore, we propose to evaluate the performance when we vary the adapter size in each stage. The results are reported in table~\ref{AblationTable_AdaptersSizes}. 

\begin{table}[t]
\centering
\caption{Effect of adapters compression rate $\sigma_i$ on adapters performance in terms for top-1 accuracy (\%). Compression rate $\sigma = \infty$ is equivalent to not adding an adapter. We vary the $\sigma_i$ value for each ViT stage (I, II, III, IV).}
\label{AblationTable_AdaptersSizes}
\resizebox{\columnwidth}{!}{
  \begin{tabular}{cccc|cc}
  \toprule
     \multicolumn{4}{c}{$\boldsymbol{\sigma}_i$ in each Swin Stage\ }&\multicolumn{2}{c}{\textbf{Dataset}} \\
    \textbf{I}& \textbf{II} & \textbf{III} & \textbf{IV}& \textbf{CIFAR-10} & \textbf{VGG-Flowers}\\
    \midrule
    ~128~& ~128~& 32& 32 & \bf 97.91 & \bf 89.90 \\
    32& 32& ~128~& ~128~ & 97.64 & 87.05 \\
    128& 32& 128& 32 & 97.84 & 89.45 \\
    32& 128& 32& 128 & 97.72 & 88.49 \\
    128& 128& 128& 32 & 97.79 & 89.22 \\
    256& 128& 64& 32 & \bf 97.91 & 89.84 \\
    32& 64& 128& 256 & 97.43 & 86.90 \\
    128& 128& $\infty$& $\infty$ & 95.29 & 73.38 \\
    $\infty$& $\infty$& 128& 128 & 97.40 & 87.04 \\
    $\infty$& 128& 128& $\infty$ & 96.97 & 84.65 \\
    128& $\infty$& $\infty$& 128 & 96.53 & 80.84 \\
    \bottomrule
 \end{tabular}%
 }
\end{table}

First, it shows that the size of adapters has a strong effect on the model performance. In general, the best performances are achieved when using adapters with a higher number of parameters. Furthermore, bigger sizes of adapters are not sufficient for better performance, but subject to which stage they are injected into. 

We observe that adding adapters to late stages (\ie~ III and IV) boosts the performance better than injecting them into early stages: adapters with $\sigma_i=128$ added to (III, IV) stages rather than (I, II) improve the performance from $95.29\%$, $73.38\%$ to $97.40\%$, $87.04\%$ on CIFAR-10 and VGG-Flowers, respectively.

\section{Illustrations of Local versus Global Neuron Removal}
\begin{figure*}[t]
\centering
\begin{subfigure}{0.33\textwidth}
  \centering
    \includegraphics[trim={0 0 450 0}, clip, width=\textwidth ]{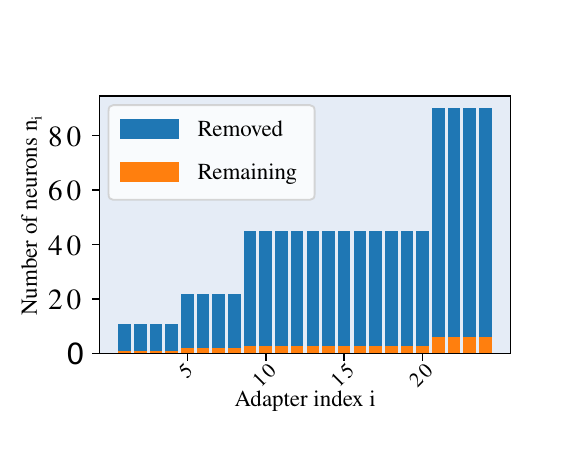}
  \caption{Local pruning: all datasets.}
\end{subfigure}%
\begin{subfigure}{0.33\textwidth}
  \centering
  \includegraphics[trim={250 0 220 0}, clip, width=0.92\textwidth]{Figures/TINA_epoch_399.pdf}
 \caption{Global pruning: VGG-Flowers .}
\end{subfigure}
\begin{subfigure}{0.33\textwidth}
  \centering
  \includegraphics[trim={505 0 0 0}, clip, width=0.78\textwidth]{Figures/TINA_epoch_399.pdf}
 \caption{Global pruning: CIFAR-10.}
    
\end{subfigure}
\caption{Layer-wise analysis of adapter's neurons distribution at $3^{rd}$ cycle. Bar plots represent a number of neurons $n_i$ at each adapter $i$ using local, and global neuron selection for VGG-Flowers and CIFAR-10, respectively.}
\label{fig:globa_vs_local399}
\end{figure*}

\begin{figure*}[t]
\centering
\begin{subfigure}{0.33\textwidth}
  \centering
    \includegraphics[trim={0 0 470 0}, clip, width=\textwidth ]{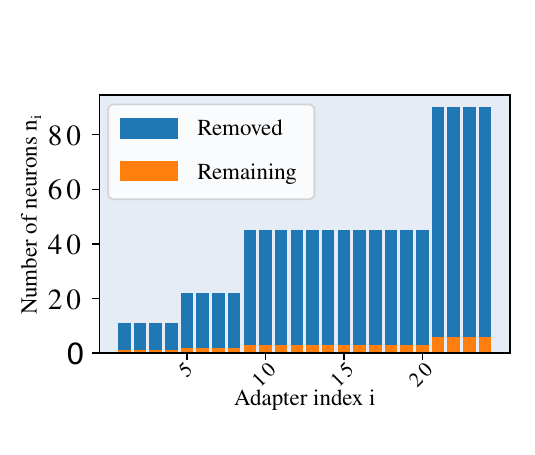}
  \caption{Local pruning: all datasets.}
\end{subfigure}%
\begin{subfigure}{0.33\textwidth}
  \centering
  \includegraphics[trim={250 0 240 0}, clip, width=0.90\textwidth]{Figures/TINA_epoch_599.pdf}
 \caption{Global pruning: VGG-Flowers.}
\end{subfigure}
\begin{subfigure}{0.33\textwidth}
  \centering
  \includegraphics[trim={495 0 0 0}, clip, width=0.88\textwidth]{Figures/TINA_epoch_599.pdf}
 \caption{Global pruning: CIFAR-10.}
\end{subfigure}
\caption{Layer-wise analysis of adapter's neurons distribution at $5^{th}$ cycle. Bar plots represent the number of neurons $n_i$ at each adapter $i$ using local and global neuron selection for VGG-Flowers and CIFAR-10, respectively.}
\label{fig:globa_vs_local599}
\end{figure*}

\begin{figure*}[t]
\centering
\begin{subfigure}{0.33\textwidth}
  \centering
    \includegraphics[trim={0 0 480 0}, clip, width=\textwidth]{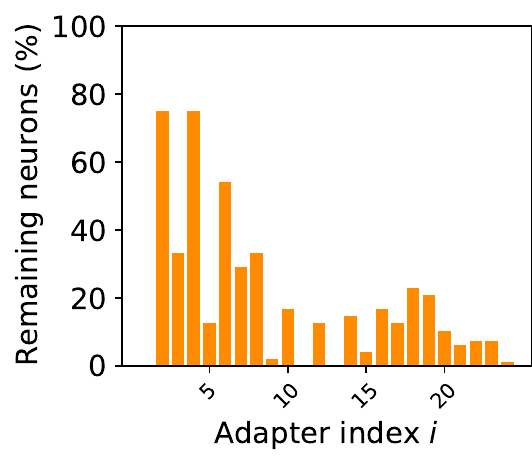}
  \caption{Local pruning: all datasets.}
\end{subfigure}%
\begin{subfigure}{0.33\textwidth}
  \centering
  \includegraphics[trim={260 0 230 0}, clip, width=0.95\textwidth]{Figures/TINA_epoch_399normalized.pdf}
 \caption{Global pruning: VGG-Flowers .}
\end{subfigure}
\begin{subfigure}{0.33\textwidth}
  \centering
  \includegraphics[trim={520 0 0 0}, clip, width=0.82\textwidth]{Figures/TINA_epoch_399normalized.pdf}
 \caption{Global pruning: CIFAR-10.}
    
\end{subfigure}
\caption{Layer-wise analysis of adapter's neurons distribution at $3^{rd}$ cycle. \textbf{Normalized} bar plots represent \textbf{percentage (\%) of remaining neurons $n_i$} at each adapter $i$ using local, global neuron selection for VGG-Flowers and CIFAR-10, respectively.}
\label{fig:norm_globa_vs_local399}
\end{figure*}

\begin{figure*}[t]
\centering
\begin{subfigure}{0.33\textwidth}
  \centering
    \includegraphics[trim={0 0 480 0}, clip, width=\textwidth ]{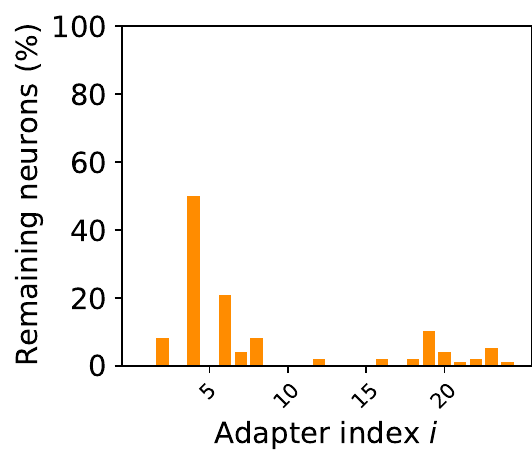}
  \caption{Local pruning: all datasets.}
\end{subfigure}%
\begin{subfigure}{0.33\textwidth}
  \centering
  \includegraphics[trim={260 0 230 0}, clip, width=0.95\textwidth]{Figures/TINA_epoch_599normalized.pdf}
 \caption{Global pruning: VGG-Flowers .}
\end{subfigure}
\begin{subfigure}{0.33\textwidth}
  \centering
  \includegraphics[trim={520 0 0 0}, clip, width=0.82\textwidth]{Figures/TINA_epoch_599normalized.pdf}
 \caption{Global pruning: CIFAR-10.}
\end{subfigure}
\caption{Layer-wise analysis of adapter's neurons distribution at $5^{th}$ cycle. \textbf{Normalized} bar plots represent \textbf{percentage (\%) of remaining neurons $n_i$} at each adapter $i$ using local, global neuron selection for VGG-Flowers and CIFAR-10, respectively.}
\label{fig:norm_globa_vs_local599}
\end{figure*}

\review{Figures~\ref{fig:globa_vs_local399} and \ref{fig:globa_vs_local599} show additional illustrations of the distribution of the removed and remaining neurons using \method{} on VGG-Flowers, and CIFAR-10. We show the learned adapters at different cycles for both local and global neuron selection methods. We also complete these visualizations (Figures \ref{fig:norm_globa_vs_local399} and \ref{fig:norm_globa_vs_local599}) with histograms where we show the percentages of remaining neurons.}
Overall, these experiments show that our method is capable of obtaining different parameter allocations that are specific to every task.



\section{Magnitude assumption as importance score}
\label{sec:appendiximportance}
In this section, we provide more insights on the importance score employed within \method{}. In particular, under Gaussian input assumption for adapters and imposing weight decay at training time, we will see that, towards a better choice of parameters to be removed, considering just $\boldsymbol{W}^{down}_i$ is sub-optimal, and $\boldsymbol{W}^{up}_i$ should be accounted as well. 
We drop the adapter index $i$ for abuse of notation, as we will always refer to the same adapter.\\
Let us have an $m$-dimensional input $\boldsymbol{h}$, whose elements are distributed according to a Gaussian $\mathcal{N}(\mu_k, \Sigma_k)$. We assume the adapter has already been trained; hence we consider, in the down-sampling phase all the $w_{jk}^{down}$ as constants. From the property of linear expectations, we know that, before reaching the non-linear activation, the post-synaptic potential is still a Gaussian random variable, having an average 
\begin{equation}
    \mu_j^{down} = \sum_{k=1}^{m} W_{jk}^{down} \cdot \mu_{k}
\end{equation}
and variance
\begin{equation}
    \label{eq:suppsigmadown}
    \Sigma_j^{down} = \sum_{k=1}^{m} W_{jk}^{down} \cdot \left[ W_{jk}^{down} \Sigma_{kk} + 2\sum_{k' < k} W_{jk'}^{down}\Sigma_{kk'} \right],
\end{equation}
where $\Sigma_{ab}$ indicates an element of the covariance matrix for the input of the adapter. For the sake of tractability, if we assume $\Sigma_{kk'}=0~\forall k\neq k'$, \eqref{eq:suppsigmadown} simply reduces to
\begin{equation}
    \Sigma_j^{down} = \sum_{k=1}^{m} \left(W_{jk}^{down}\right)^2  \Sigma_{kk}.
\end{equation}
In transformers, the commonly-used activation function is the Gaussian error linear unit (GELU), whose analytical expression is
\begin{equation}
    \phi(x) = x\cdot \frac{1}{2}\left[ 1+\text{erf}\left(\frac{x}{\sqrt{2}}\right)\right]
\end{equation}
where erf$(\cdot)$ is the error function. For values close to zero, or larger than zero, it can be approximated to the identity function, while for values much lower than zero, it asymptotically tends to zero. Let us focus on the first scenario: we can approximate the post-synaptic potential to the output of the non-linearity, saying that the output
\begin{equation}
    z_{j}\approx \mathcal{N}(\mu_j^{down}, \Sigma_j^{down}).
\end{equation}
At this point, the signal undergoes an up-sampling: following up on the same approach adopted for the down-sampling, we find that the output $\boldsymbol{r}$ still follows a Gaussian distribution having an average
\begin{equation}
    \mu_l^{up} = \sum_{j=1}^{n} W_{jl}^{up}, \mu_j^{down} = \sum_{j=1}^{n} W_{jl}^{up} \sum_{k=1}^{m} W_{jk}^{down} \cdot \mu_{k}
\end{equation}
and variance
\begin{equation}
    \Sigma_l^{up} = \sum_{j=1}^{n} \left(W_{jl}^{up}\right)^2 
    \end{equation}
    \begin{equation}
    \Sigma_{j}^{down} = \sum_{j=1}^{n} \left(W_{jl}^{up}\right)^2 \sum_{k=1}^{m} \left(W_{jk}^{down}\right)^2  \Sigma_{kk}
\end{equation}

\begin{align}
    \mu_{l,\bar{a}}^{up} &= \mu_l^{up} - W_{al}^{up} \sum_{k=1}^{m} W_{ak}^{down} \cdot \mu_{k}\nonumber\\
    \Sigma_{l, \bar{a}}^{up} &= \Sigma_l^{up} - \left(W_{al}^{up}\right)^2 \sum_{k=1}^{m} \left(W_{ak}^{down}\right)^2  \Sigma_{kk}
\label{eq::stdbar}
\end{align}

In order to assess the impact of removing a whole neuron in the embedding space, we can write the KL-divergence of the distribution for $r_l$ with and without the $a$-th neuron in the embedding space:
\begin{multline}
\label{eq:KLprelim}
 \KL(r_l,r_{l,\bar{a}}) =\\
 \log\left(\frac{\Sigma_l^{up}\!-\!\left(W_{al}^{up}\right)^2 \sum_{k=1}^{m} \left(W_{ak}^{down}\right)^2  \Sigma_{kk}}{\Sigma_l^{up}}\right) \nonumber \\ 
 +\frac{\left(\Sigma_l^{up}\right)^2+\left(\mu_l^{up} - \mu_l^{up} + W_{al}^{up} \sum_{k=1}^{m} W_{ak}^{down} \cdot \mu_{k} \right)}{2 \cdot \left[ \Sigma_l^{up} - \left(W_{al}^{up}\right)^2 \sum_{k=1}^{m} \left(W_{ak}^{down}\right)^2  \Sigma_{kk}\right]^2} - \frac{1}{2}.\tag{12}
\end{multline}

According to \eqref{eq::stdbar}, we rewrite \eqref{eq:KLprelim} as \ref{eq:KL2}.

\vspace{-4mm}
\begin{multline}
\label{eq:KL2}
 \KL(r_l,r_{l,\bar{a}}) = \\
 \log\left(1\!-\!\frac{\left(W_{al}^{up}\right)^2 \sum_{k=1}^{m} \left(W_{ak}^{down}\right)^2  \Sigma_{kk}}
 {\Sigma_l^{up}}\right)\nonumber\\
 + \frac{\left(\Sigma_l^{up}\right)^2+\left(W_{al}^{up} \sum_{k=1}^{m} W_{ak}^{down} \cdot \mu_{k} \right)}{2 \cdot \left[ \Sigma_l^{up} - \left(W_{al}^{up}\right)^2 \sum_{k=1}^{m} \left(W_{ak}^{down}\right)^2  \Sigma_{kk}\right]^2} - \frac{1}{2}.\tag{13}
\end{multline}

Let us now investigate which is the $a$-th neuron which, when removed, causes the least perturbation at the output $r_l$ (or in other words, such that $\KL(r_l,r_{l,\bar{a}})$ is as low as possible). Looking at the argument of the logarithm, we ask $\frac{\left(W_{al}^{up}\right)^2 \sum_{k=1}^{m} \left(W_{ak}^{down}\right)^2  \Sigma_{kk}}{\Sigma_l^{up}} = 0$ and, since we can safely assume $\Sigma_l^{up}$, we need to select $a$ such that $\left(W_{al}^{up}\right)^2 \sum_{k=1}^{m} \left(W_{ak}^{down}\right)^2  \Sigma_{kk} = 0$. Considering that also $\Sigma_{kk}> 0~\forall k$, we satisfy the condition if either:\begin{itemize}
    \item $W_{ak}^{down}=0~\forall k$, namely the $\normlone$ norm for $\boldsymbol{W}_{-a}^{down}$ is zero;
    \item $W_{al}^{up}=0$. Considering though that this condition needs to be satisfied for all the $l$ outputs of the adapters, we ask $W_{al}^{up}=0~\forall l$ or, in other words, the $\normlone$ norm for $\boldsymbol{W}_{a-}^{up}$ is also zero.
\end{itemize}
We observe that, when either of the two conditions is met, the KL divergence is zero as
\begin{equation*}
    \KL(r_l,r_{l,\bar{a}}) = \log(1) + \frac{\left(\Sigma_l^{up}\right)^2}{2\cdot \left(\Sigma_l^{up}\right)^2} - \frac{1}{2} = 0
\end{equation*}
We can also assume that if either $W_{ak}^{down}=0~\forall k$ or $W_{al}^{up}=0~\forall l$, the norm of the non-zero parameters associated with some neuron $a$ are small when training with any weight penalty regularizer (as the contribution to the output is zero, the signal is either not forward or back-propagated, leaving the weight penalty term the only update for these parameters). 

\section{Detailed evaluation of \method{} on MultiTask/DomainNet benchmarks}
\label{DetailsEvaluationBenchmarks}
Table~\ref{Domainnettable} reports the number of trained parameters and the average accuracy across datasets in the DomainNet benchmark. For both, the number of trained parameters is reported in millions, and the average top-1 accuracy on the datasets is reported in the rightest column.

We observe that \emph{full fine-tuning} has generally the highest accuracy, but it requires a huge number of parameters to be finetuned for each dataset. Among the vanilla fine-tuning baselines, we observe that tuning the parameters of the \emph{attention/MLP} layer turns out to be surprisingly effective. Nevertheless, it still requires a high number of task-specific parameters, compared to other PET approaches. \emph{Linear probing} does not perform well illustrating the need to change the feature representations of the model when adapting to new tasks.

\emph{PHM}, and \emph{Compacter} are effective methods to get on-par performance with \emph{full-model finetune} while adjusting less than 2\% of the parameters. Contrarily to what is observed for NLP tasks \cite{EfficientNLP}, PETs on visual tasks do not reach \emph{full fine-tuning} performance on any dataset \review{with a low number of trainable parameters (smaller than 2\%)}. \review{VPT does not perform well, indicating that injecting tokens into the embedding space does not help much if the pre-training dataset is different from the downstream task}. Generally speaking, all PET methods maintain similar performance rankings on all tasks. This suggests that the choice of the best adaptation strategy does not depend on the downstream task.

\noindent\emph{Adapters} outperform all PET methods in terms of accuracy ($69.39\%$ for DomainNet, $92.91\%$ for Multi-task) but just with a higher number of trainable parameters (1.37M, $4.90\%$ of the total) for $\sigma=32$.

\begin{figure}[t]
\begin{center}
\includegraphics[trim={5 25 5 40}, clip, width=\linewidth]{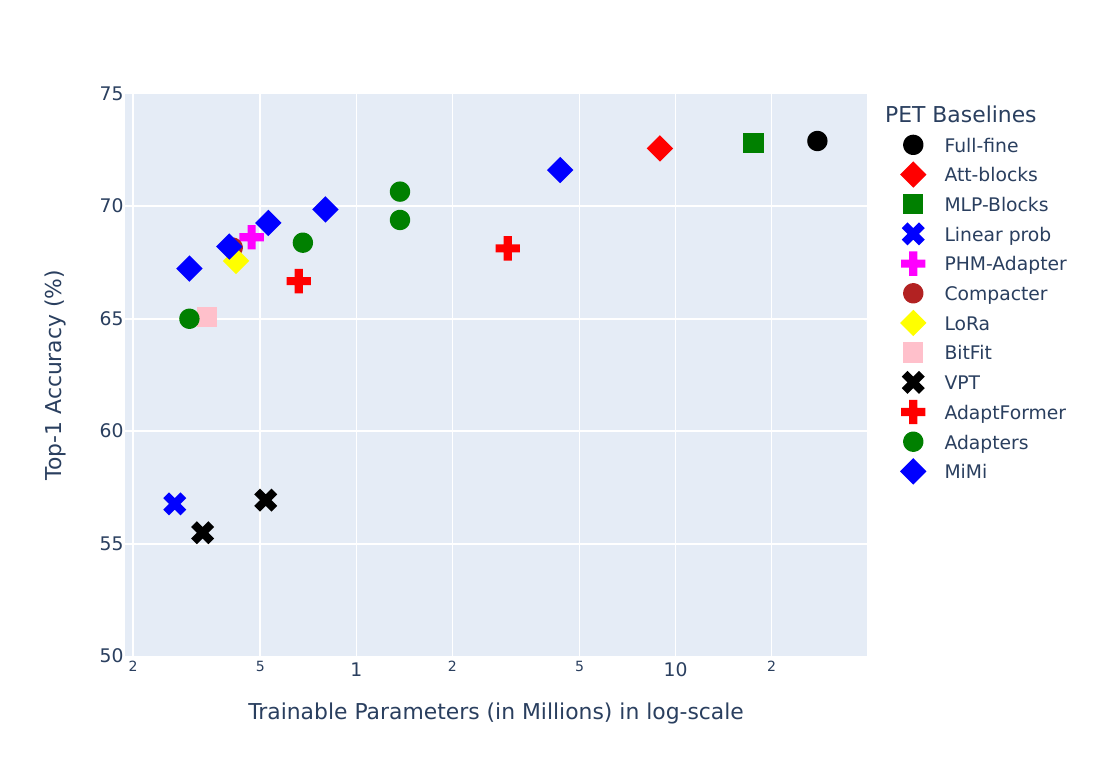}
\end{center}
   \caption{Evaluation of PET baselines mean top-1 accuracy on \textbf{DomainNet} benchmark. Reducing \method{} (\textcolor{blue}{\ding{117}}) parameters does not significantly reduce performance compared to other PET baselines.}
\label{fig:PETs_evaluation_domainnet}
\end{figure}

\emph{Adapters} outperform \emph{AdaptFormer} with fewer parameters (92.91\% with 1.37M parameters, versus 92.27\% with 2.98M parameters). This result indicates that adapting the representations after both MSA and MLP blocks, as done in \emph{Adapters} (see Fig. \ref{fig:PETs_evaluation_domainnet}), allows better adaptation than acting only on the MLP block via a parallel branch (as done in \emph{AdaptFormer}~\cite{AdapterFormer}).

When comparing \emph{adapter} with uniform and proportional parameter distribution, we observe that allocating parameters proportionally to the layer dimension performs better. Indeed, adapters with $\sigma=32$ outperform adapters with $n_{i}=47\forall i$ ($70.65\%$ vs $69.39\%$ in DomainNet, $93.53\%$ vs $92.91\%$ in Multi-task). This suggests that the last layers, which have higher dimensionality, are more task-specific, and consequently require more adaptation. We also show that reducing the size of adapters ($n_{i}=23$) hurts the performance with a drop which, despite being marginal for Multi-task ($0.23\%$) is more consistent in DomainNet ($1.01\%$). This emphasizes that training tiny adapters in a vanilla fashion leads to unsatisfying performance and motivates our specific training procedure.

\begin{table*}[t]
  \centering
  \caption{Results on the DomainNet benchmark~\cite{DomainNet}. C, I, P, Q, R, and S stand for Clipart, Infograph, Painting, Quickdraw, Real, and Sketch respectively. $^\dag$ is equivalent to Adapters with $\sigma =8$. Methods are grouped according to the relative number of trainable parameters (\colorbox{tiny}{$\leq 2\%$}, \colorbox{medium}{$\in ]2,10[\%$},\colorbox{large}{$\geq 10\%$}) 
}
\label{Domainnettable}
  \resizebox{0.75\linewidth}{!}{
  \begin{tabular}{lcc|ccccccc}
    \toprule 
	\textbf{Method} & \# \textbf{Params}& \textbf{Trained}& \textbf{C.} & \textbf{I.} & \textbf{P.} & \textbf{Q.} & \textbf{R.} & \textbf{S.} & \textbf{Mean} \\
    & \textbf{(M)} $\downarrow$&\textbf{(\%)} $\downarrow$&&&&&&&$\uparrow$ \\
	\midrule 
	Full fine-tuning & \colorbox{large}{27.8} & \colorbox{large}{100} & 
	
	79.16& 48.29& 74.64& 75.88& 86.21& 73.26 & 
	\textbf{72.90}\\ 
	Att-blocks & \colorbox{large}{8.93} & \colorbox{large}{32.14} & 
    48.36& 75.38&	73.28&	86.13&	72.81 & 72.57&72.57\\ 
	MLP-blocks & \colorbox{large}{17.54} & \colorbox{large}{63.12}& 
	79.23& 48.11& 75.02& 74.82& 86.35& 73.29 & 72.80\\ 

 \textbf{\method{}} (0 cycle)$^\dag$ & \colorbox{large}{4.44} & \colorbox{large}{16.15}& 
	78.11& 46.93& 74.38& 72.19& 86.09& 71.97 & 71.61 \\ %
     \midrule
    \review{AdaptFormer-256} & \colorbox{medium}{2.54} & \colorbox{medium}{9.24}&
    \review{75.32}& \review{43.74}& \review{72.16}& \review{66.00}& \review{83.88}& \review{67.68} & \review{68.13} \\
    \review{AdaptFormer-64} & \colorbox{medium}{0.84} & \colorbox{medium}{3.06} &
	\review{73.76}& \review{42.38}& \review{71.11}& \review{63.41}& \review{83.23}& \review{66.14} & \review{66.67}\\
    VPT (100 tokens) & \colorbox{medium}{0.71} & \colorbox{medium}{2.57} & 64.33 & 18.65 & 63.20 & 59.40 & 79.65 & 56.41 &  56.94\\

 \review{Adapters} ($\sigma =32$)& \colorbox{medium}{1.37} & \colorbox{medium}{4.90} & 77.42&	46.51& 74.06 &	69.81&	85.30&	70.84 & \textbf{70.65}  \\ 
    Adapters ($n_i=47$)& \colorbox{medium}{1.37} & \colorbox{medium}{4.90} &  76.15 &	45.28 & 73.04 &	67.86 &	84.83 &	69.17 & 69.39  \\ 

    Adapters ($n_i=23$)& \colorbox{medium}{0.68} & \colorbox{medium}{2.47} & 
    75.28& 44.17& 72.41& 66.44& 83.98& 68.02 & 68.38  \\ 
    
	\textbf{\method{}} (1 cycle) & \colorbox{medium}{0.80} & \colorbox{medium}{2.89}& 
	76.83& 46.00& 73.76& 67.93& 85.05& 69.61 & 69.86 \\ 
    
    \midrule
 Linear prob & \colorbox{tiny}{\textbf{0.27}} & \colorbox{tiny}{\textbf{0.95}} &
	62.89& 33.96& 64.93& 42.95& 81.69& 54.24& 56.77 \\ 
    
PHM-Adapter & \colorbox{tiny}{0.47} &  \colorbox{tiny}{1.72}&
	
	75.79 & 44.62 & 72.49 & 66.62 & 83.73 & 68.51& 68.62\\  
	
	Compacter & \colorbox{tiny}{0.41} & \colorbox{tiny}{1.44}& 
	75.25 & 44.19 & 72.09 & 66.01 & 83.42 & 67.99 & 68.16 \\ 
	
	
	BitFit & \colorbox{tiny}{0.34} & \colorbox{tiny}{1.22}&
	
	72.14& 41.07& 70.00& 60.39& 82.43& 64.43&  65.08 \\
	
	VPT (10 tokens) & \colorbox{tiny}{0.32} & \colorbox{tiny}{1.15}&
	61.91& 24.87& 57.05 & 57.09 & 76.94 & 55.12 &  55.49\\

        \imad{SSF} & \colorbox{tiny}{0.28} & \colorbox{tiny}{0.96}&
	74.45 & 42.64 & 71.72 & 64.70 & 82.95 & 65.90 &  67.06 \\
        
        \imad{Fact-TK$_{32}$} & \colorbox{tiny}{0.33} & \colorbox{tiny}{1.18}&
	72.46 & 41.14 & 70.32 & 61.26 & 80.23 & 63.87 &  64.88 \\
    
        \review{Adapters ($n_i=1$)} & \colorbox{tiny}{0.30} &\colorbox{tiny}{{1.07}} & 
        \review{72.12}& \review{41.30}& \review{69.93}& \review{59.95}& \review{82.49}& \review{64.18}& \review{65.00} \\

	\textbf{\method{}} (2 cycles) & \colorbox{tiny}{0.53} &\colorbox{tiny}{1.92}&
	76.83& 45.45& 73.11& 66.67& 84.42& 69.05 & \textbf{69.26} \\ 

	\textbf{\method{}} (3 cycles) & \colorbox{tiny}{0.40} & \colorbox{tiny}{1.43} &
	75.44& 44.60& 72.59& 64.73& 83.87& 68.05 &68.21\\
	
	\textbf{\method{}} (4 cycles) & \colorbox{tiny}{0.30} &\colorbox{tiny}{1.07} & 
	74.38& 43.52& 71.50& 63.41& 83.12 & 67.46 &67.23\\
	\bottomrule
    \end{tabular}
    }
\end{table*}

\subsection{Impact of $\rho$}
\label{sec:rho}
In this section, we investigate the effect of the hyper-parameter $\rho$ (amount of neuron removal) in our method \method{}.

\begin{figure}[t]
    \centering
    \includegraphics[trim={10 10 10 50}, clip, width=\linewidth]{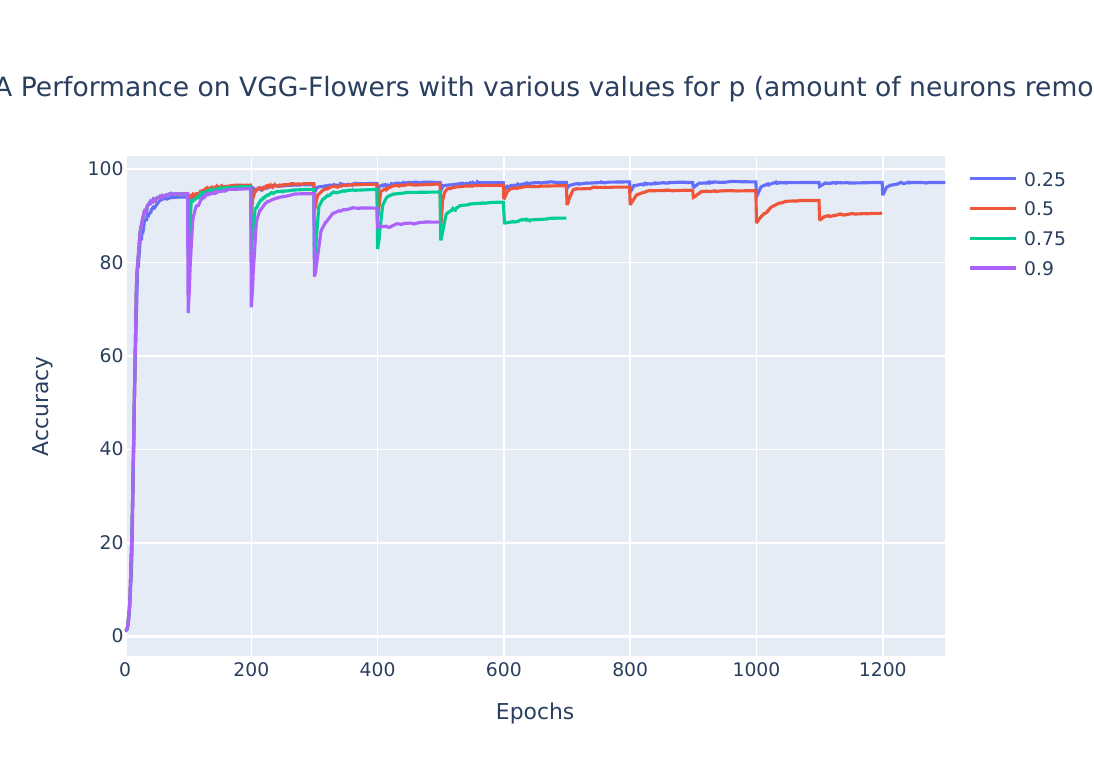}
    \caption{Analysis of \method{} performance on VGG-Flowers dataset with different values of $\rho$. If $\rho$ is very high, the drop in performance is significant, but it requires less $C$ training cycles to reach $\sigma_{traget}$.}
    \label{fig:rho_analysis}
\end{figure}

In Fig.~\ref{fig:rho_analysis}. We notice that higher values of $\rho$ hurt the performance because we remove many parameters after each cycle, but we reduce the size of adapters significantly. On the other hand, if $\rho$ is small (i.e 25\%), we maintain good performance on the VGG-Flowers dataset, but it requires higher training cycles $C$ to reach the target compression rate $\sigma_{target}$.

We have a trade-off between the performance, and training budget in order to reach the $\sigma_{target}$. Removing too many parameters at each cycle hurts performance. Maintaining good performance requires a higher number of training cycles $C$.

\subsection{Evaluation on VTAB Benchmark}
\label{sec:VTAB_results}
We experiment on \textit{VTAB}~\cite{zhai2019vtab} is a collection of 19 diverse visual classification tasks, which are organized into three groups: 
\textit{Natural} - tasks that contain natural images captured using standard cameras;
\textit{Specialized} - tasks that contain images captured via specialized equipment, such as medical and satellite imagery;
and \textit{Structured} - tasks that require geometric comprehension like object counting.
Each task of VTAB contains 1000 training examples. Following~\cite{zhai2019vtab}, we use the provided 800-200 split of the train set to determine hyperparameters and run the final evaluation using the full training data. We report the average accuracy score on the test set within three runs.

\begin{landscape}
  \setlength{\textwidth}{8.875in}
  \setlength{\textheight}{6.875in}
\begin{table*}
\caption{Per-task fine-tuning results for VTAB with a pre-trained ViT-B/16. Results for other baselines are reported from~\cite{VPT}.}
\label{table:supp_vtab}
\resizebox{\textwidth}{!}{   
\begin{tabular}{
c l 
rrrrrrr r
rrrr r
rrrrrrrr r}
\toprule
  &&\rotatebox{90}{\bf{CIFAR-100}}
  &\rotatebox{90}{\bf{Caltech101} }
  &\rotatebox{90}{\bf{DTD} }
  &\rotatebox{90}{\bf{Flowers102} }
  &\rotatebox{90}{\bf{Pets} }
  &\rotatebox{90}{\bf{SVHN} }
  &\rotatebox{90}{\bf{Sun397} }
  &\rotatebox{90}{\bf{Mean}}
  &\rotatebox{90}{\bf{Patch Camelyon} }
  &\rotatebox{90}{\bf{EuroSAT} }
  &\rotatebox{90}{\bf{Resisc45} }
  &\rotatebox{90}{\bf{Retinopathy} }
  &\rotatebox{90}{\bf{Mean}}
  &\rotatebox{90}{\bf{Clevr/count} }
  &\rotatebox{90}{\bf{Clevr/distance} }
  &\rotatebox{90}{\bf{DMLab}}
  &\rotatebox{90}{\bf{KITTI/distance} }
  &\rotatebox{90}{\bf{dSprites/location} }
  &\rotatebox{90}{\bf{dSprites/orientation} }
  &\rotatebox{90}{\bf{SmallNORB/azimuth} }
  &\rotatebox{90}{\bf{SmallNORB/elevation} }
  &\rotatebox{90}{\bf{Mean}}
  \\
\midrule
\bf{(a)}&\fullft{} &68.9 &87.7 &64.3 &97.2 &86.9 &87.4 &38.8 &75.88 &79.7 &95.7 &84.2 &73.9 &83.36 &56.3 &58.6 &41.7 &65.5 &57.5 &46.7 &25.7 &29.1 &47.64 
\\
\midrule
\multicolumn{4}{ l }{\emph{Head-oriented}} 
\\
\multirow{6}{*}{\bf{(a)}}
&\linear{} &63.4 &85.0 &63.2 &97.0 &86.3 &36.6 &51.0 &68.93 (1) &78.5 &87.5 &68.6 &74.0 &77.16 (1) &34.3 &30.6 &33.2 &55.4 &12.5 &20.0 &9.6 &19.2 &26.84 (0)
\\
&\partialft{}-1 &66.8 &85.9 &62.5 &97.3 &85.5 &37.6 &50.6 &69.44 (2) &78.6 &89.8 &72.5 &73.3 &78.53 (0) &41.5 &34.3 &33.9 &61.0 &31.3 &32.8 &16.3 &22.4 &34.17 (0)
\\
&\mlp{}-2 &63.2 &84.8 &60.5 &97.6 &85.9 &34.1 &47.8 &67.70 (2) &74.3 &88.8 &67.1 &73.2 &75.86 (0) &45.2 &31.6 &31.8 &55.7 &30.9 &24.6 &16.6 &23.3 &32.47 (0)
\\
&\mlp{}-3 &63.8 &84.7 &62.3 &97.4 &84.7 &32.5 &49.2 &67.80 (2) &77.0 &88.0 &70.2 &56.1 &72.83 (0) &47.8 &32.8 &32.3 &58.1 &12.9 &21.2 &15.2 &24.8 &30.62 (0)
\\
&\mlp{}-5 &59.3 &84.4 &59.9 &96.1 &84.4 &30.9 &46.8 &65.98 (1) &73.7 &87.2 &64.8 &71.5 &74.31 (0) &50.8 &32.3 &31.5 &56.4 &7.5 &20.8 &14.4 &20.4 &29.23 (0)
\\
&\mlp{}-9 &53.1 &80.5 &53.9 &95.1 &82.6 &24.4 &43.7 &61.90 (1) &78.5 &83.0 &60.2 &72.3 &73.49 (0) &47.5 &27.9 &28.9 &54.0 &6.2 &17.7 &10.8 &16.2 &26.15 (0)
\\
\midrule
\multicolumn{4}{ l }{\emph{Backbone-oriented}} 
\\
\multirow{6}{*}{\bf{(b)}}
&\sidetune{} &60.7 &60.8 &53.6 &95.5 &66.7 &34.9 &35.3 &58.21 (0) &58.5 &87.7 &65.2 &61.0 &68.12 (0) &27.6 &22.6 &31.3 &51.7 &8.2 &14.4 &9.8 &21.8 &23.41 (0)
\\
&\bias{} &72.8 &87.0 &59.2 &97.5 &85.3 &59.9 &51.4 &73.30 (3) &78.7 &91.6 &72.9 &69.8 &78.25 (0) &61.5 &55.6 &32.4 &55.9 &66.6 &40.0 &15.7 &25.1 &44.09 (2)
\\
&\adapter{}-256 &74.1 &86.1 &63.2 &97.7 &87.0 &34.6 &50.8 &70.50 (4) &76.3 &88.0 &73.1 &70.5 &76.98 (0) &45.7 &37.4 &31.2 &53.2 &30.3 &25.4 &13.8 &22.1 &32.39 (0)
\\
&\adapter{}-64 &74.2 &85.8 &62.7 &97.6 &87.2 &36.3 &50.9 &70.65 (4) &76.3 &87.5 &73.7 &70.9 &77.10 (0) &42.9 &39.9 &30.4 &54.5 &31.9 &25.6 &13.5 &21.4 &32.51 (0)
\\
&\adapter{}-8 &74.2 &85.7 &62.7 &97.8 &87.2 &36.4 &50.7 &70.67 (4) &76.9 &89.2 &73.5 &71.6 &77.80 (0) &45.2 &41.8 &31.1 &56.4 &30.4 &24.6 &13.2 &22.0 &33.09 (0)
\\
\midrule
\multicolumn{4}{ l }{\emph{Visual-Prompt Tuning}} 
\\
\multirow{6}{*}{\bf{(c)}}&shallow-VPT &77.7 &86.9 &62.6 &97.5 &87.3 &74.5 &51.2 &76.81 (4) &78.2 &92.0 &75.6 &72.9 &79.66 (0) &50.5 &58.6 &40.5 &67.1 &68.7 &36.1 &20.2 &34.1 &46.98 (4)
 \\
 &Prompt length ($p$) &100 &5 &1 &200 &50 &200 &1 &79.4 &5 &50 &50 &10 &28.7 &100 &200 &100 &100 &100 &100 &200 &200 &137.5
 \\
&Tuned / Total (\%) &0.18 &0.10 &0.04 &0.27 &0.08 &0.19 &0.36 &0.17 &0.01 &0.05 &0.09 &0.01 &0.04 &0.10 &0.18 &0.09 &0.09 &0.10 &0.10 &0.19 &0.19 &0.13
\\
\cmidrule{2-24}
&deep-VPT &78.8 &90.8 &65.8 &98.0 &88.3 &78.1 &49.6 &78.48 &81.8 &96.1 &83.4 &68.4 &82.43 &68.5 &60.0 &46.5 &72.8 &73.6 &47.9 &32.9 &37.8 &54.98
\\
&Prompt length ($p$) &10 &10 &10 &1 &1 &50 &5 &12.4 &100 &100 &10 &1 &52.8 &50 &200 &100 &50 &10 &50 &200 &200 &107.5
\\
&Tuned / Total (\%) &0.20 &0.20 &0.15 &0.10 &0.04 &0.54 &0.41 &0.23 &1.06 &1.07 &0.15 &0.02 &0.57 &0.54 &2.11 &1.07 &0.54 &0.12 &0.55 &2.12 &2.11 &1.14 
\\
\midrule
\multirow{1}{*}{\bf{(Ours)}}& \method{} &61.39 &86.77 &66.65 & 96.13 &90.98 & 79.3 & 53.4 & 76.37 & 83.06 & 95.77 & 85.9 & 75.51 & 85.06 & 62.57 & 65.77 & 46.43 & 74.91 & 76.58 & 53.57 & 24.59 & 35.91 & 55.04 \\
\bottomrule
\end{tabular}
}
\end{table*}
\end{landscape}

\section{Details about Datasets}

In Table~\ref{datasets}, we report different statistics that capture the diversity of the datasets we use in our experiments.

\begin{table}[ht]
\centering
      \centering
      \begin{tabular}{lcccr}
        \toprule
        
        \multicolumn{1}{c}{} & \textbf{Dataset} & \textbf{Train Size} & \textbf{Test Size} & \textbf{Classes} \\
        \midrule
        
        \multirow{4}*{\rotatebox{90}{Multi-task}}
        & CIFAR-10 & 50000 &	10000& 10 \\
        & CIFAR-100 & 50000 &	10000& 100 \\
        & Oxford Flowers &2040 &6149 &102 \\
        & SVHN &  73257 & 26032 & 10 \\
        
        \midrule
        \multirow{6}*{\rotatebox{90}{DomainNet}}
        & Clipart    & 33525 & 14604 & 345 \\
        & Infograph & 36023 & 15582 & 345 \\
        & Painting & 50416 & 21850 & 345  \\
        & Quickdraw & 120750 & 51750 & 345 \\
        & Real & 120906 & 52041 & 345 \\
        & Sketch & 48212 & 20916 & 345 \\
        \bottomrule
        \end{tabular}
        \caption{Datasets used in our empirical analysis}
      \label{datasets}
\end{table}

\newpage

\setlength{\tabcolsep}{4pt}
\begin{table*}[ht]
\small
\begin{center}
\caption{Specifications of the various datasets evaluated. 
$^{\star}$: we randomly sampled the \texttt{train} and \texttt{val} sets since there are no public splits available
}
\label{table:supp_datasets}
\resizebox{0.95\linewidth}{!}{%
\begin{tabular}{l l  l l l l l}
\toprule
\textbf{Dataset}   &\textbf{Description}  & \textbf{\# Classes}    &\textbf{Train}  &\textbf{Val}  &\textbf{Test} \\ 
\midrule
\multicolumn{3}{l}{Fine-grained visual recognition tasks (FGVC)} 
\\
\cmidrule{2-6}
\quad\cub{}~\cite{WahCUB_200_2011}
& Fine-grained bird species recognition
&200
&5,394$^{\star}$	&600$^{\star}$ &5,794	
\\

\quad\nabirds{}~\cite{van2015nabirds}
& Fine-grained bird species recognition
&55
&21,536$^{\star}$	&2,393$^{\star}$	&24,633
\\

\quad\flowers{}~\cite{nilsback2008automated}
& Fine-grained flower species recognition
&102
&1,020	&1,020	&6,149 
\\

\quad\dogs{}~\cite{Khosla_FGVC2011dogs}
 &Fine-grained dog species recognition  &120 
 &10,800$^{\star}$	&1,200$^{\star}$	&8,580 
\\

\quad\cars{}~\cite{gebru2017cars}
& Fine-grained car classification  &196  
&7,329$^{\star}$	&815$^{\star}$	&8,041 
\\

\midrule

\multicolumn{3}{l}{Visual Task Adaptation Benchmark (\vtab{})~\cite{zhai2019vtab}} 
\\
\cmidrule{2-6}
\quad CIFAR-100~\cite{cifar10} &\multirow{7}{*}{Natural}
&100 &\multirow{7}{*}{800/1000} &\multirow{7}{*}{200} &10,000
\\
  \quad Caltech101~\cite{li2006one} & &102 && &6,084
  \\
  \quad DTD~\cite{cimpoi14describing} & &47 && &1,880
  \\
  \quad Flowers102~\cite{nilsback2008automated} & &102 && &6,149
  \\
  \quad Pets~\cite{parkhi12a} & &37 && &3,669
  \\
  \quad SVHN~\cite{netzer2011reading} & &10 && &26,032
  \\
  \quad Sun397~\cite{xiao2010sun} & &397 && &21,750
  \\
\cmidrule{2-6}

  \quad Patch Camelyon~\cite{veeling2018rotation} &\multirow{4}{*}{Specialized} &2
  &\multirow{4}{*}{800/1000} &\multirow{4}{*}{200} &32,768
  \\
  \quad EuroSAT~\cite{helber2017eurosat} & &10 && &5,400
  \\
  \quad Resisc45~\cite{cheng2017remote} & &45 && &6,300
  \\
  \quad Retinopathy~\cite{kaggle-diabetic-retinopathy} & &5 && &42,670
  \\

\cmidrule{2-6}
  \quad Clevr/count~\cite{johnson2017clevr} &\multirow{8}{*}{Structured}
  &8
  &\multirow{8}{*}{800/1000} &\multirow{8}{*}{200} &15,000
  \\
  \quad Clevr/distance~\cite{johnson2017clevr} & &6 && &15,000
  \\
  \quad DMLab~\cite{beattie2016deepmind} & &6 && &22,735
  \\
  \quad KITTI/distance~\cite{Geiger2013IJRR} & &4 && &711
  \\
  \quad dSprites/location~\cite{dsprites17} & &16 && &73,728
  \\
  \quad dSprites/orientation~\cite{dsprites17} & &16 && &73,728
  \\
  \quad SmallNORB/azimuth~\cite{lecun2004learning} & &18 && &12,150
  \\
  \quad SmallNORB/elevation~\cite{lecun2004learning} & &9 && &12,150
\\

\bottomrule\end{tabular}
}
\end{center}
\end{table*}
\setlength{\tabcolsep}{1.4pt}




\end{document}